\documentclass{article}

\usepackage{PRIMEarxiv}

\usepackage[utf8]{inputenc} % allow utf-8 input
\usepackage[T1]{fontenc}    % use 8-bit T1 fonts
\usepackage{hyperref}       % hyperlinks
\usepackage{url}            % simple URL typesetting
\usepackage{booktabs}       % professional-quality tables
\usepackage{amsfonts}       % blackboard math symbols
\usepackage{nicefrac}       % compact symbols for 1/2, etc.
\usepackage{microtype}      % microtypography
\usepackage{lipsum}
\usepackage{fancyhdr}       % header
\usepackage{graphicx}       % graphics
\graphicspath{{media/}}     % organize your images and other figures under media/ folder
\usepackage{authblk}
\usepackage{amsmath} 
\usepackage{chapterbib}

\title{Landslide topology uncovers failure movements
}

\author[1,2,3,$\dag$,*]{Kamal Rana}
\author[4,$\dag$,*]{Kushanav Bhuyan}
\author[3,5]{Joaquin Vicente Ferrer}
\author[2,6]{Fabrice Cotton}

\author[2,3]{Ugur Ozturk}
\author[4]{Filippo Catani}
\author[7]{Nishant Malik}

\affil[1]{{Chester F. Carlson Center for Imaging Science}, {Rochester Institute of Technology, Rochester, 14623, NY, USA}}

\affil[2]{Helmholtz Centre Potsdam – GFZ German Research Centre for Geosciences, Potsdam, 14473, Brandenburg, Germany}

\affil[3]{Institute of Environmental Science and Geography, University of Potsdam, Potsdam, 14473, Brandenburg, Germany}

\affil[4]{{Machine Intelligence and Slope Stability Laboratory, Department of Geosciences, University of Padova, Padua, 35129, Veneto, Italy}}

\affil[5] {{Potsdam Institute for Climate Impact Research, Potsdam, 14473, Brandenburg, Germany}}

\affil[6]{Institute of Geosciences, University of Potsdam, Potsdam, 14473, Brandenburg, Germany}

\affil[7]{School of Mathematical Sciences, Rochester Institute of Technology, Rochester, 14623, NY, USA}

\affil[$\dag$]{These authors contributed equally and share first authorship.}

\begin{document}
\maketitle

\begin{abstract}
The death toll and monetary damages from landslides continue to rise despite advancements in predictive modeling. The predictive capability of these models is limited as landslide databases used in training and assessing the models often have crucial information missing, such as underlying failure {types}. Here, we present an approach for identifying failure {types based on their movements, e.g., slides and flows} by leveraging 3D landslide topology. We observe topological proxies reveal prevalent signatures of mass movement mechanics embedded in the landslides’ morphology or shape, {such as detecting coupled movement styles within complex landslides.} We find identical failure types exhibit similar topological properties, and by using them as predictors, we {can} identify failure {types} {in historic and event-specific landslide databases (including multi-temporal) from various geomorphological and climatic contexts} such as Italy, the US’s Pacific Northwest region, {Denmark, Turkey, and China} with 80-94\% accuracy. {To demonstrate the real-world application of the method, we implement it in two undocumented datasets from China and publicly release the datasets.} These new insights can considerably improve the performance of landslide predictive models and impact assessments. Moreover, our work introduces a new paradigm for studying landslide shapes to understand underlying processes through the lens of landslide topology.
\end{abstract}

% keywords can be removed

\section{Introduction}
Eery year, landslides cause economic damages worth $20$ billion US dollars {\cite{klose2015landslide}}, and between $2004$ and $2019$ non-seismic landslides alone caused about $70,000$ fatalities worldwide \cite{froude2018global}. Within the first two months of $2023$, we have seen reports of devastating landslides in São Paulo, Brazil \cite{reuters2023}, Southern Peru \cite{aljaze2023}, and New Zealand \cite{reuters2023newzealand}, injuring many and killing approximately 70 people. %The current landslide crisis in Joshimath, India, is also worsening, with signs of subsidence owing to creep moments inflicting damage to  $\sim$$800$ buildings and endangering $181$ people living atop the landslide. 
Adding to this, recent studies count over one million landslide occurrences with annual volumes estimated at fifty-six billion cubic {meters} globally \cite{broeckx2020landslide}, presenting a risk to sixty million people \cite{ozturk2022climate}. With the increase in urbanization, global climate change, and environmental change trends, the frequency of landslides and the associated risks will keep increasing globally over time \cite{ozturk2022climate}. In line with this, landslides are anticipated to evolve and remobilize with increased frequency under changing climatic conditions on a decadal scale \cite{fan2021rapidly, gariano2016landslides}. Our ability to identify hazards from emerging landslides and dynamically assess impact areas is essential in averting risk to rapidly urbanizing communities and adapting to changing environmental conditions \cite{lima2023conventional, ozturk2022climate}.
%Moreover, with the increase in urbanization, global climate change, and environmental change trends, the frequency of landslides and the associated risks will keep increasing globally with time \cite{ozturk2022climate}. 

\begin{figure*}[t!]
\includegraphics[width=\columnwidth]{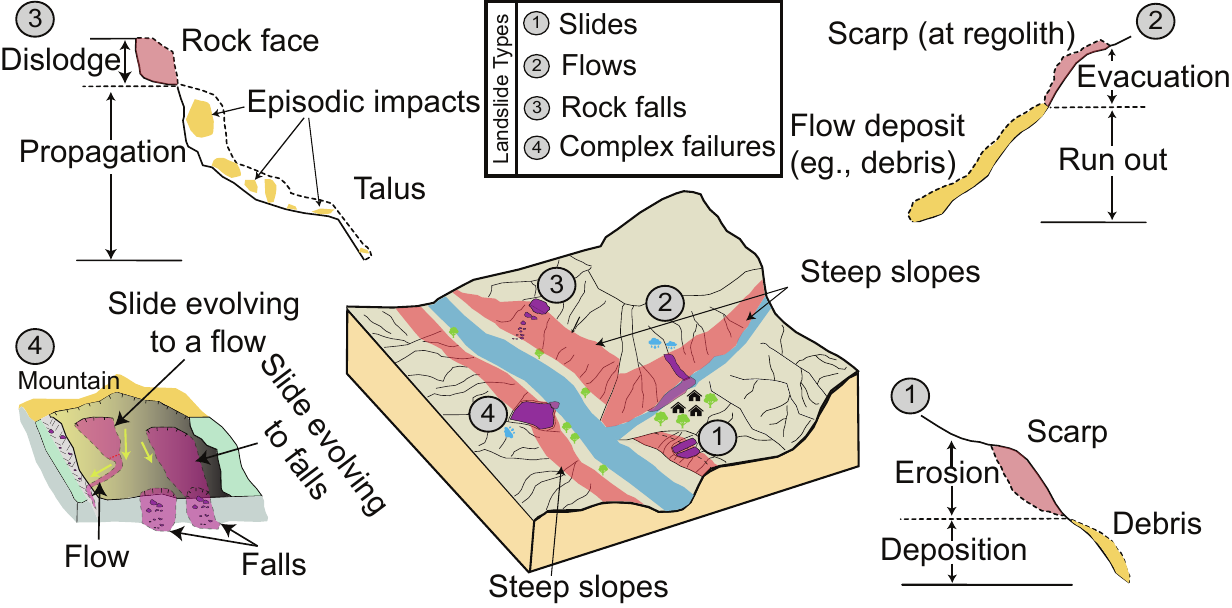}\hfill
\caption{The schematic {broadly} illustrates different landslide failure {types} and their associated mechanical and kinematic behavior{, excluding sub-type movements}. For example, Type 1 refers to slide-type failures that can constitute deep ruptures, where a cohesive unit of soil or rock slides down a slope following a well-defined rupture plane. Type 2 refers to flow-type failures where soil, rock, or other material {travels} down a slope as a dense, fluid-like mass with a flow-like motion. Type 3 is a fall-type of failure where a body of rock detaches from a steep slope or cliffs and exhibits free falling and episodic impacts as it propagates down the slope. Type 4 refers to the complex interaction and effect of numerous geomorphic processes transpiring in a single failure event, where processes start as one {types} and evolve into another; such as a slide-type failure evolving to a flow-type failure.}
\label{fig:conceptual}
\end{figure*}

To address the rising landslide risk, predictive models for hazard, risk, and early warning systems are developed which assist in forecasting landslide occurrences and locating landslide-prone regions to mitigate the associated impacts \cite{corominas2014recommendations}. However, the efficacy of these models is contingent on the quality of the underlying landslide databases. These databases often lack the much-needed information about the {type of failure} of the mapped landslides \cite{guzzetti2012landslide}. 
% Baffling predictive models (e.g., Huang et al.\cite{huang2023uncertainties}), these databases instead rely on a broader definition of landslides that covers all types of gravitational mass wasting processes, such as slides, flows, and falls {(including sub-types based on movement, e.g., rotational slides \cite{hungr2014varnes})}.
{Generally, these databases include a broader definition of landslides that covers all types of gravitational mass wasting processes, such as slides, flows, and falls including sub-types based on their movement, e.g., rotational slides \cite{hungr2014varnes} combined together, thereby hampering the capability of predictive models.}
Typically, each landslide failure {type} exhibits different geological, geometrical, and geotechnical properties (see Figure~\ref{fig:conceptual}). For instance, slides have conspicuous primary scarps and collapse along the planar or rotational surfaces \cite{varnes1978slope}, flows such as mudflows exhibit visco-plastic or viscous/fluid kinematics caused by excess pore water pressure \cite{bradley2019earthquake}, and rock falls entails the free falling of fragmented rocks from steep slopes \cite{bourrier2013use} (see Supplementary Section S2 for detailed explanations). Practitioners usually combine these different failure {movements} into one group within an inventory, despite their different properties \cite{guzzetti2006estimating, lombardo2018presenting, rossi2010optimal}, since categorizing them manually requires comprehensive surveys (remote and field) and standardized classification protocols \cite{guzzetti2012landslide}, which are laborious and time-consuming. Consequently, predictive models start to harbor significant levels of uncertainty and bias \cite{huang2023uncertainties}, hence failing to match empirical observations, especially when moving from local levels to regional and global scales \cite{kirschbaum2010global,reichenbach2018review,fressard2014data}. 
% {For instance, a possible implication of this might be that predictive models identify landslide occurrences with inaccurate locations, timings, and magnitudes.} %, while also mischaracterizing the involved type-specific failures.}

% For instance, they may predict a low probability of landslide occurrence in a high landslide-prone region. Therefore, identifying landslide failure types {in a time-efficient manner} is crucial to improving predictive modeling.

Preliminary attempts at identifying failure {movement types} have considered both knowledge-driven and data-driven approaches. While the former are region-specific, bounded by expert-based rules, and constrained to small areas \cite{martha2010characterising,Barlow2006687}, the latter addressed these problems with supervised learning and have successfully identified landslide failure types in the Italian context \cite{amato2021data}. However, the existing solutions are still limited in their prediction capabilities, as the failure information is derived from geometric properties of two-dimensional (2D) landslide polygons (outlining the landslide planforms). Owing to the inherent limitations of 2D landslide polygons, crucial kinematic and mechanical details embedded in the landslides' three-dimensional (3D) morphology are overlooked, such as the style of kinematic progression, deformation patterns and structures, and debris deposits. Furthermore, the {complex} kinematic evolution of one or more failure types may culminate in the convergent evolution of landslide shapes, wherein landslides starting as completely different movements may evolve to follow similar planar outlines. This presents challenges in deducing failure {types} solely based on 2D geometric descriptors or rudimentary topographic metrics. We argue that these overlooked kinematic attributes are intricately linked with the 3D morphological properties of landslides, which can be comprehensively analyzed through topological methods.
% This phenomenon complicates the discernment of {failure information} based solely on 2D representations or simple topographic measures. We posit that such morphological and kinematic information is rooted in the landslide’s 3D morphology which can be extracted via topology. 

Topology is a sub-discipline of mathematics explored in many fields that concern the study of shapes \cite{lum2013extracting}, such as in protein structures, data modeling, complex networks, and signal processing \cite{luo2023sensing, carlsson2020topological, han2022topological, zangeneh2019topological}. We explore advanced data analysis tools rooted in topology, known as topological data analysis (TDA), which captures critical structures present in the data’s shape (in our case, the landslide’s 3D shape). We hypothesize that key features of landslide kinematics are embedded in the 3D topology of the involved landforms and that TDA properties can capture their kinematic movements as a proxy for identifying landslide failures.

% We hypothesize that TDA properties/features can capture the kinematics and mechanics as a proxy for identifying landslide failure mechanisms. %After extracting topological properties from landslide 3D shapes, we feed them into a supervised machine-learning algorithm–--random forest---to determine failure mechanisms.

\begin{figure*}[t!]
\includegraphics[width=\columnwidth]{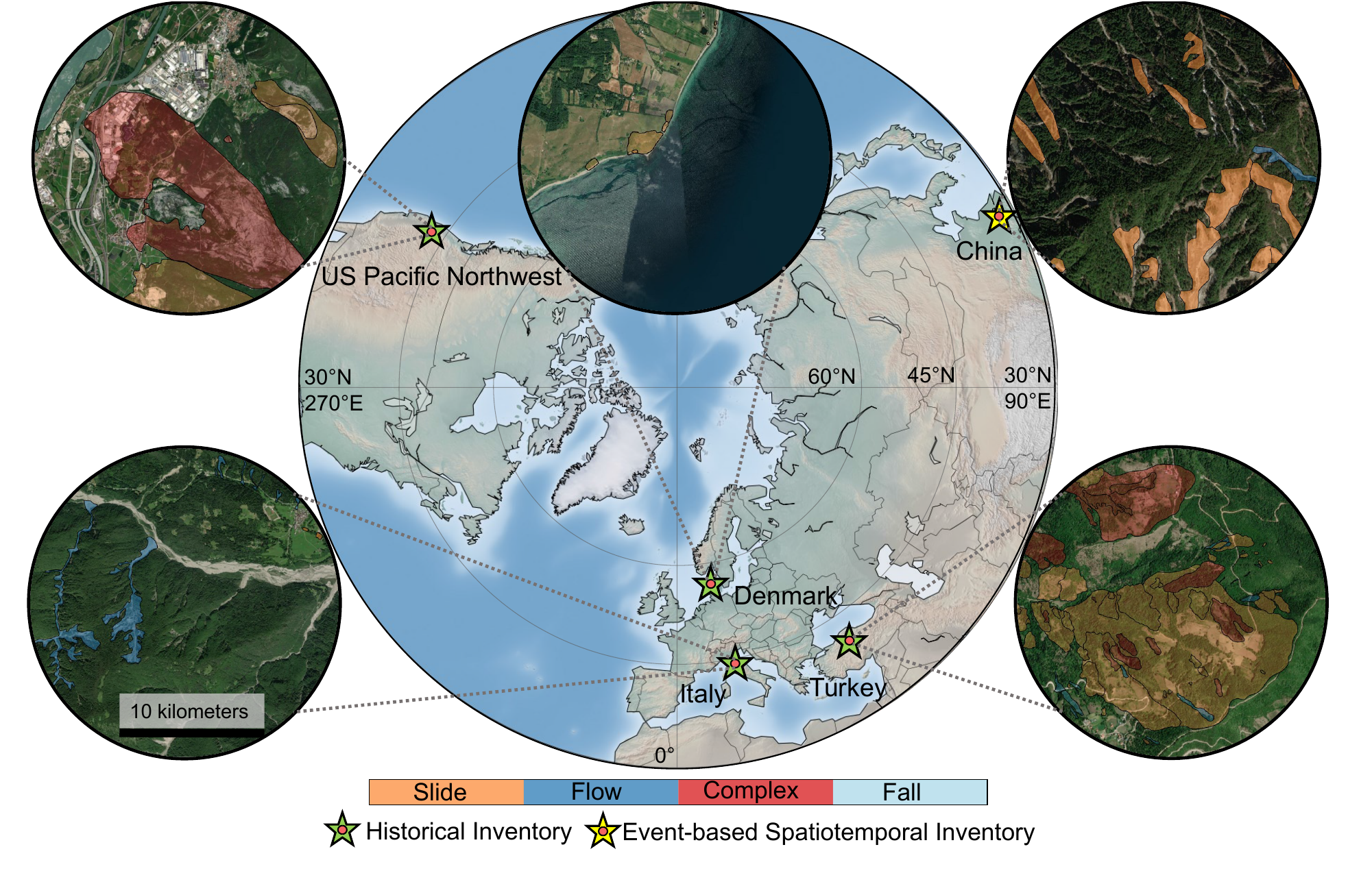}\hfill
\caption{The diagram shows the geographical regions: {Italy, the US Pacific Northwest, Turkey, Denmark, and China} whose data we analyzed in this work. The inset in the circular shape shows snippets from diverse regions with landslide polygons of different failure movements on top of the World Imagery from ESRI. Map credits: Esri, Maxar, Earthstar Geographics, and the GIS User Community \cite{ESRI2020}}
\label{fig:datamap}
\end{figure*}

In this study, we introduce an approach for uncovering landslide failure {types based on their mode of movement} by examining the topological properties inherent in the 3D shapes of landslides. {We develop the method using the Italian historical inventory from IFFI and then deploy it to landslide inventories from varying geomorphological and climatic settings: the United States (US) Pacific Northwest region, Denmark, Turkey, and China  (see Figure~\ref{fig:datamap}) to validate the effectiveness and applicability of the approach}. We demonstrate that this method offers a more comprehensive understanding of the underlying failure {types} compared to traditional analyses based on 2D polygonal geometry. {We also explore the model's ability to identify sub-types of the failure movements (e.g., translational and rotational slides, earth and debris flow). Also, we utilized the topological properties of complex landslides to reveal coupled failure types underlying the formation of complex landslides. In addition, we identify the types of failure in an event-based multi-temporal landslide inventory, enhancing our understanding of landslide dynamics and behaviors in real-world scenarios. {Moreover, we deployed our method on two undocumented (paleo-event) datasets as a real-world application, which we verified using Google Earth archive imageries.}
To the best of our knowledge, sub-type failure movements, temporal prediction of failure movements, and detection of the underlying failure types within complex landslides have never been investigated using an automated data-driven approach.} Here, we showcase with our findings that the proposed method (1) is user-friendly, exclusively requiring only the landslide polygonal shape and a Digital Elevation Model (DEM) as input, (2) exhibits high performance in discerning failure types based on their movements, (3) is transferable across various geomorphological and climatic regions, and (4) shows strong performance and remains robust, even when the availability of samples is limited, indicating its {wide} applicability in data-scarce regions. By offering a deeper understanding of landslide failure {movements}, this approach has the potential to enhance the accuracy and reliability of landslide susceptibility, hazard, and risk assessment models, by providing valuable insights to the predictive modeling community.

% In this work, we present a method to identify landslide failure mechanisms based on the topological properties of their 3D shapes. We find that landslide topology provides deeper insight into failure mechanisms than the two-dimensional (2D) polygonal geometry of landslides. To evaluate the efficacy of our approach, we implement it in the landslide databases of two geomorphologically different regions: Italy and the Pacific North-West region of the US (see Figure~\ref{fig:datamap}). Here we showcase that the present method (1) is easy-to-use as it exclusively depends on the landslide polygonal shape as input, (2) has a strong performance in identifying failure mechanisms, (3) is transferable to other geomorphological regions, and (4) robust as it achieves strong performance even with fewer training samples. The proposed approach has significant implications for improving the accuracy and reliability of landslide susceptibility, hazard, and risk assessment models, with its applications primarily focused on the predictive modeling community.

\section{Methods}
\subsection{Topological Feature Engineering}
\begin{figure*}[t!]
\includegraphics[width=\columnwidth]{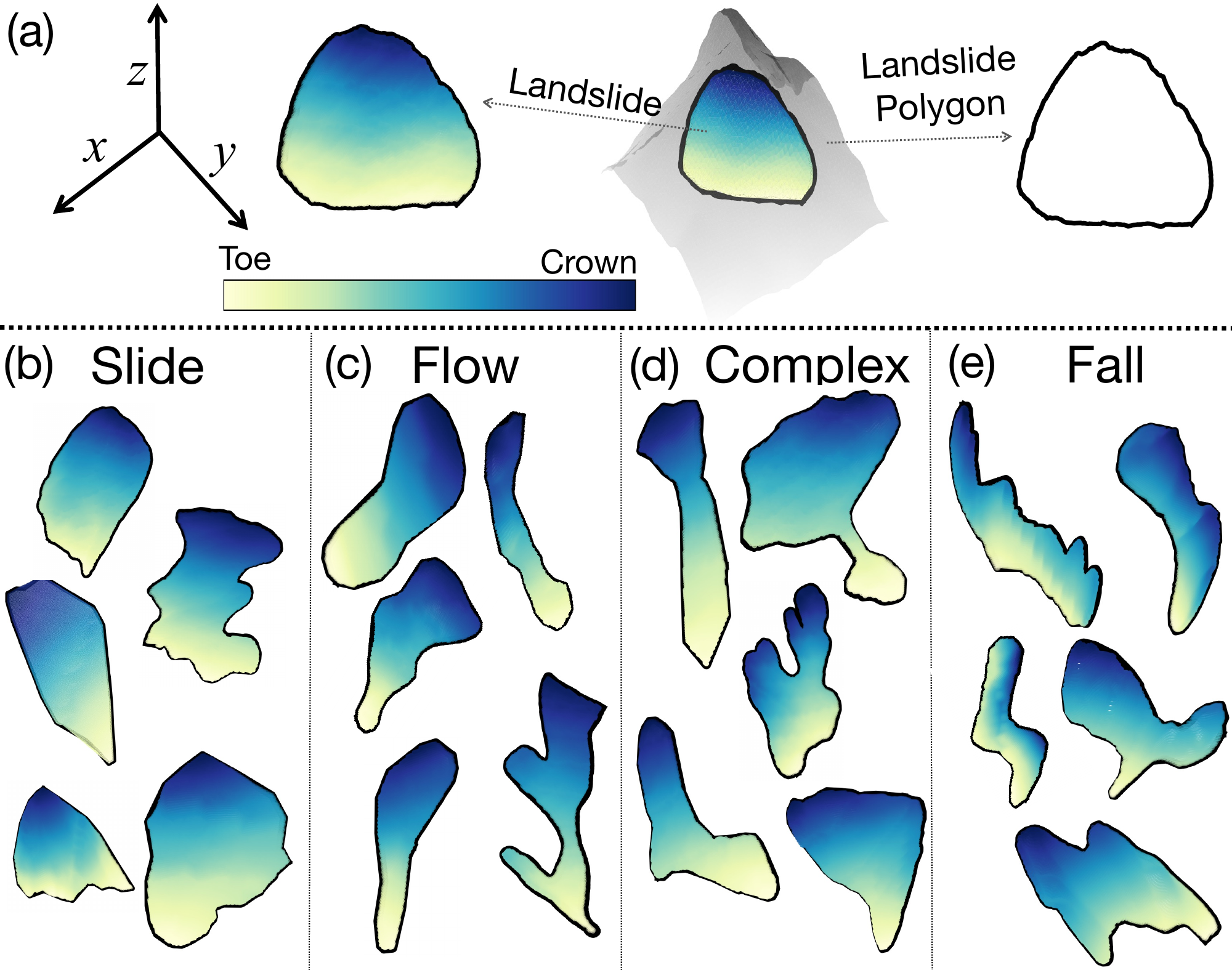}\hfill
\caption{(a) An example of a landslide failure in a terrain with a steep slope. The
diagram also shows the 3D landslide polygon, which outlines the landslide shape. (b-e) 3D landslide samples for different landslide failure types, namely, the slide, flow, complex, and fall types.}
\label{fig:landsamples}
\end{figure*}

%Polygon-based landslide databases carry crucial information regarding the occurrence date of landslides, triggering factor, volume, and the failure type, which are essential for various applications such as developing a landslide early warning system or near-real-time predictive tools (CITE) and could be used for rapid landslide hazard assessment (CITE). Landslide polygons follow the outline of the landslide body on the ground however, in the databases, they are available as 2D polygons. We converted these 2D polygons into 3D polygonal meshes where each mesh corresponded to a landslide in the 3-dimensional domain including the geographical location of each polygonal vertex. Therefore, we developed a method for finding the landslide types based entirely on the landslide polygon shapefiles as the input to the method. In the proposed method, we generated the 3D shapes using the z-values from DEMs, and then extracted topological information about the landslide's 3D shape using Topological Data Analysis (TDA). Finally, we explored various TDA features and using feature importance analysis, we fed the relevant TDA features to the machine learning algorithm for classifying landslide failure types.

\begin{figure*}[t!]
\includegraphics[width=\columnwidth]{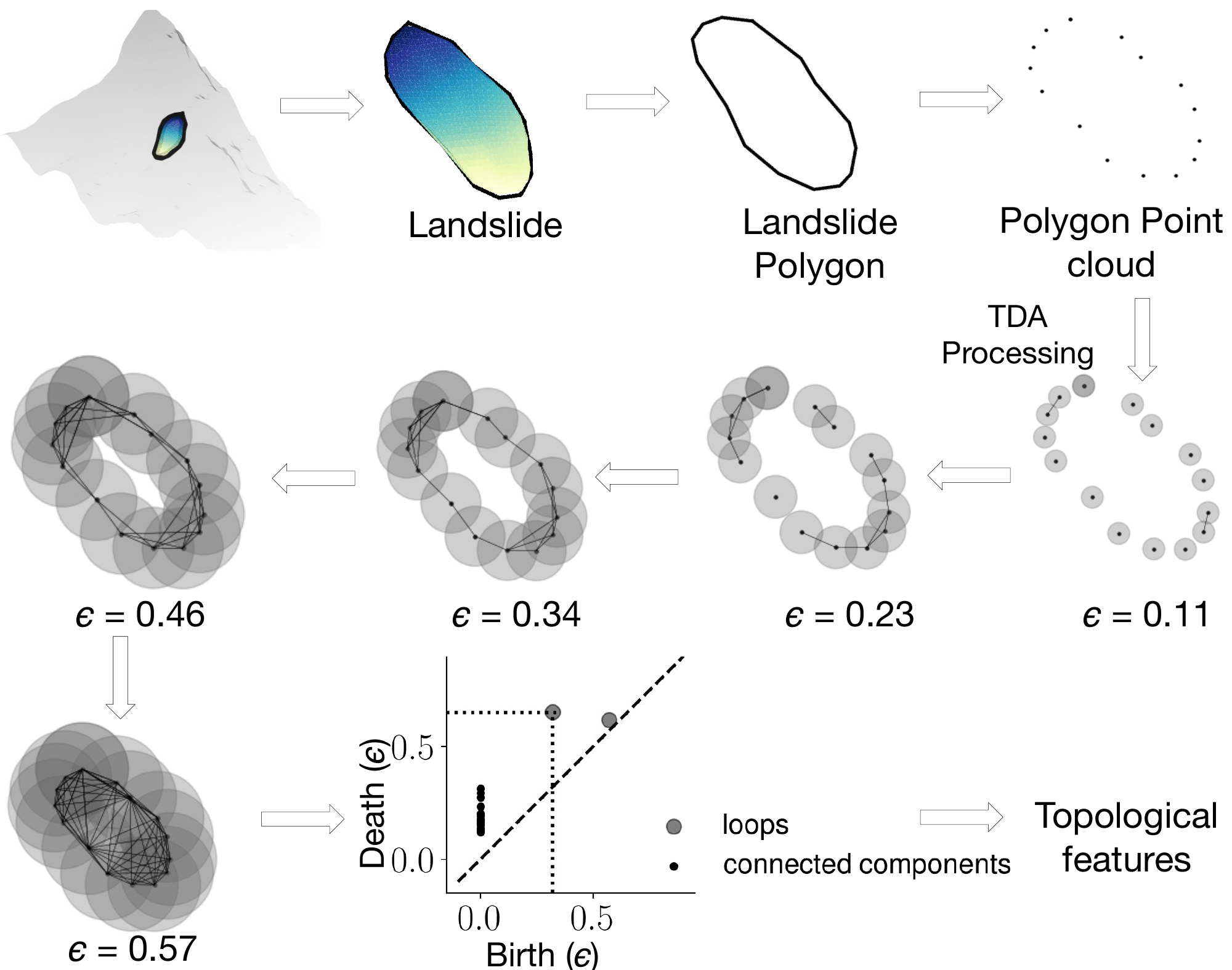}\hfill
\caption{The diagram illustrates the procedure of computing topological features corresponding to a landslide shape. The flowchart shows the use of persistent homology in capturing various structures of the landslide shape by using an evolving disk size ($\epsilon$) around each point in the point cloud. With the increase in $\epsilon$, various structures like connected components and holes emerge in the data’s shape which is captured by the persistence diagram. Using this information, we can calculate the topological properties of the landslide's shape. Please note that when processing the TDA features, we display the flowchart using a 2D illustration for simplicity and better visualization.}

% The persistence diagram captures the birth and death of structures (such as holes and connected components), and using this information, we can calculate the topological properties of the landslide's shape.

\label{fig:datafig1}
\end{figure*}

In the proposed method, landslide polygons serve as the primary input. These polygons represent the 2D outline of the landslide body on the ground and are commonly found in landslide databases. Each vertex of the landslide polygon comprises geographical latitude and longitude coordinates. Utilizing the Digital Elevation Model, the landslide polygons are transformed into normalized 3D shape outlines, wherein each vertex encompasses latitude, longitude, and elevation information. Topological Data Analysis (TDA) is employed to extract the geometrical and topological characteristics of a landslide's 3D shape outline (see Figure~\ref{fig:datafig1}). This information is subsequently used as input for a machine learning algorithm, specifically the random forest. The Python library Giotto-TDA is leveraged to extract an assortment of TDA properties/features from the 3D shape of landslides \cite{tauzin2021giotto}. To ascertain the most pertinent features for landslide-type classification, a correlation test is conducted between TDA features, and those with high correlation are removed. The remaining, less correlated features are then assessed, and the least important ones are iteratively eliminated until six robust predictors remain. The exclusion of additional predictors results in decreased performance, while incorporating more than seven yields comparable outcomes. Utilizing fewer predictors facilitates the development of a more generalizable model. The six features thus form a feature space for the random forest classifier.

Topological Data Analysis (TDA) quantifies the multidimensional shape of data using algebraic topology techniques. TDA offers a variety of metrics for capturing the geometric and topological properties of data shape \cite{carlsson2009topology}. These metrics could be used as a feature space for the machine learning algorithm to solve various classification and regression problems, such as shape classification. TDA's central idea is persistent homology, which identifies persistent geometric features by using simplicial complexes to extract topological features from point cloud data. Simplicial complexes are a collection of simplexes that are the building blocks of higher-dimensional counterparts of a graph. An n-dimensional simplex is formed by connecting n+1 affinely independent points \cite{munch2017user,garin2019topological}. For example, a point is a 0-dimensional simplex, an edge that connects two points is a 1-dimensional simplex, and a filled triangle formed by combining three non-linear points is a 2-dimensional simplex. A Vietoris-Rips complex indicates the simplicial complex in the data's shape using a parameter $\epsilon$. The main idea of the Vietoris-Rips complex is to connect any two points in the point cloud data set whose distance is less than  $\epsilon$. These connections of data points create structures in the data that change with the parameter $\epsilon$. Therefore, to get complete information about all the structures in the data, the idea is to use all $\epsilon>0$ values.

% Only particular structures present in the data shape provide relevant information about the data shape. Homology measures these unique structures present in the data like $0$-dimensional homology captures connected components or clusters, $1$-dimensional homology measures loops, and $2$-dimensional homology measures voids. These crucial structures emerge and die with changes in  $\epsilon$, and this information is captured in the persistence diagram. With the help of a persistence diagram, we can calculate various measures quantifying the topological properties of the shape. These measures are- persistence entropy, average lifetime, number of points, betti curve-based measure, persistence landscape curve-based measure, Wasserstein amplitude, Bottleneck amplitude, Heat kernel-based measure, and landscape image-based measure. We used all these measures as an input to the machine learning method--random forest. 

Only specific structures in the data shape provide crucial information about the geometrical and topological properties of the data. Homology measures these unique structures in the data, where e.g., $0$-dimensional homology captures connected components or clusters, $1$-dimensional homology measures loops, and $2$-dimensional homology measures voids \cite{munch2017user,hensel2021survey}. These crucial structures emerge and die with changes in  $\epsilon$, and this information is captured in the persistence diagram.
With the help of a persistence diagram, we can calculate various measures quantifying the topological properties of the shape-- persistence entropy, average lifetime, number of points, Betti curve-based measure, persistence landscape curve-based measure, Wasserstein amplitude, Bottleneck amplitude, Heat kernel-based measure, and landscape image-based measure \cite{bubenik2017persistence, reininghaus2015stable, adams2017persistence}. We have explained all these topological features in detail in Supplementary Section S3. Finally, we used all these measures as input in the machine learning method--random forest.

\subsection{Machine learning model: Random Forest}
Random forest is an ensemble-based learning method that has shown promising results in various classification and regression problems \cite{barnett2019endnote,biau2016random,breiman,kursa2014robustness,chaudhary2016improved}. Random forest classifiers consist of multiple classifiers trained independently on bootstrapping training samples. Bootstrapping $N$ training samples leads to $\frac{2N}{3}$ independent samples, so each tree in the random forest is constructed from a distinct subset of training samples \cite{azar2014random,belgiu2016random}. Moreover, each tree in the random forest predicts the output class of the testing sample independently, and the class with the majority votes is the final decision of the random forest \cite{arabameri2021decision,belgiu2016random}. 

Each random forest tree divides a parent node into two daughter nodes, right ($r$) and left ($l$). For each node split, the random forest chooses $p$ features from the $m$ total features of the samples \cite{azar2014random,okun2007random}. Among $p$ features, the random forest selects a single feature for a node split based on the "Gini-index" criterion. The Gini-index for each right and left daughter node can be calculated as: $G_{r}$=1-$\Sigma_{j=1}^{j=N} P_{rj}$ and 
$G_{l}$=1-$\Sigma_{j=1}^{j=N} P_{lj}$. Here, $P_{rj} (P_{rlj})$ and $N$ are the probability of the samples in the right (left)nodes having class $j$ and the total number of the classes. The features that maximize the change in the Gini-index that is calculated as follows: ${{\displaystyle\Delta \theta (s_q)= G_q-\rho_{rq} G_r-\rho_{lq} G_l}}$ is used for the node split \cite{kuhn2013applied,zhang2012ensemble}. Here, $\rho_{rq}$ and $\rho_{lq}$ are the proportion of samples in the right and left daughter nodes. The process of splitting nodes continues until a stopping criterion is met, such as when no more samples are available for splitting, or when the Gini-index of parent nodes is lower than that of daughter nodes.

\section{Results}
\begin{figure*}[t!]
\includegraphics[width=\columnwidth]{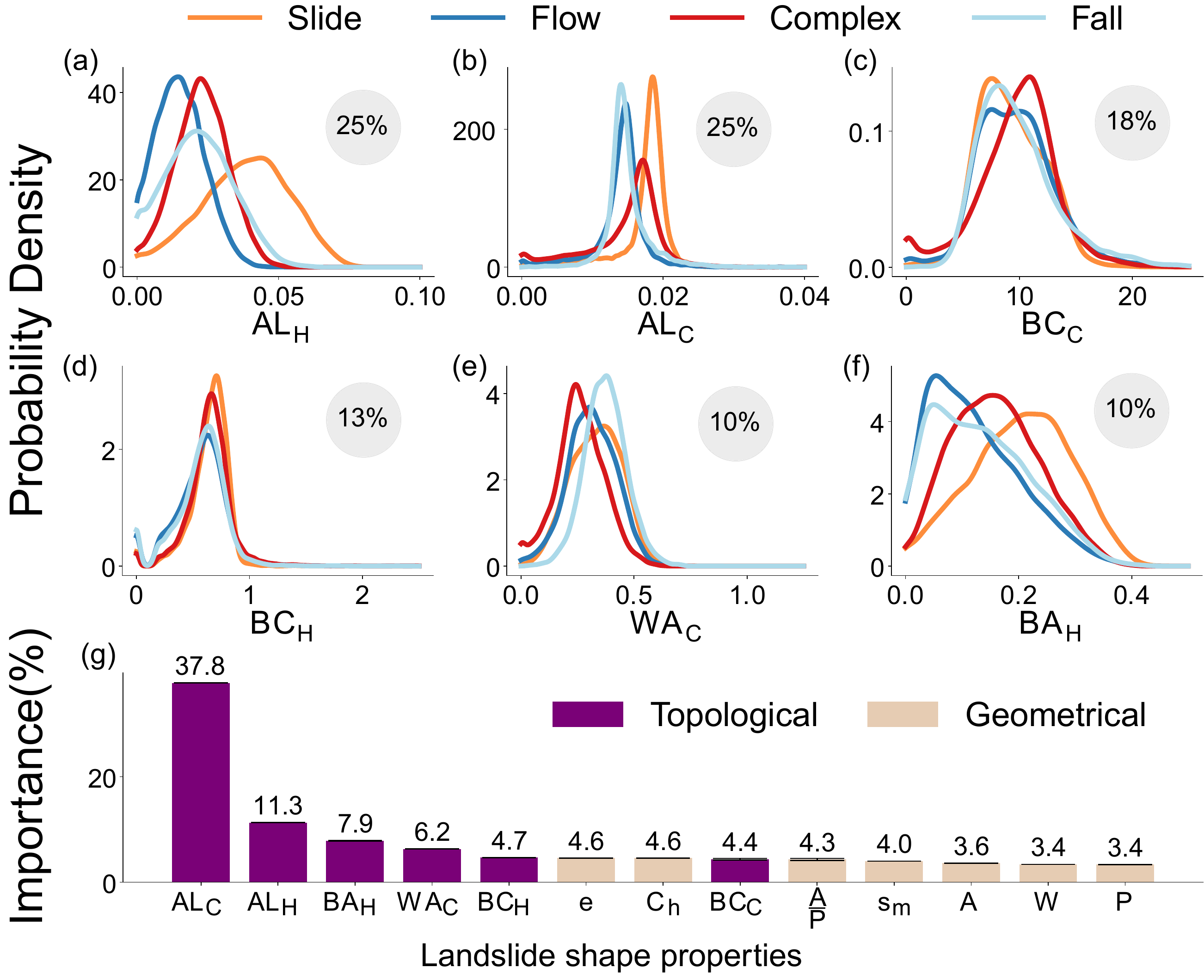}\hfill
\caption{Plots (a-f) show the probability distribution functions of the six most optimal topological properties used in classifying the failure {types} for slides, flows, complex, and falls in Italy. Please note that we discuss the probability distribution functions
of different {failure types} for the Italian region only, as the Italian data set is the most data-rich inventory. The y-axis shows the probability density values (calculated using kernel density estimation), and the x-axis shows the value of topological attributes. The topological properties in plots (a-f) are: Average lifetime of holes (AL\textsubscript{H}), Average lifetime of connected components (AL\textsubscript{C}), Betti-curve based feature of connected components (BC\textsubscript{C}), Betti-curve based feature of holes (BC\textsubscript{H}), Wasserstein amplitude of holes (WA\textsubscript{H}), and Bottleneck amplitude of holes (BA\textsubscript{H}) (the computations of these properties are explained in detail in Supplementary Section S3 and Figure~\ref{fig:datafig1}). The percentage values in the gray circular disk in each figure indicate the topological feature's importance (in \%), as estimated by the random forest-based classification procedure. Sub-plot (g) shows the joint computed feature importance of topological and geometric properties by the random forest model. The analysis shows topological properties consistently outperform geometric properties with a standard deviation under 0.1\% (The error bar represents the standard deviation. However, it is not visible in plot-g because the standard deviation is very small). The geometric properties from top to bottom are: area ($A$), perimeter ($P$), the ratio of area to perimeter $\frac{A}{P}$, convex hull-based measure ($C_h$), minor($s_m$),  and width ($W$) of the minimum area bounding box fitted to the polygon.
}
\label{fig:tdafeatures}
\end{figure*}

\subsection{ Landslide topology as a proxy to identify failure {types based on style of movement}}

The underpinning of topological data analysis (TDA) is rooted in structures in the data's shape, such as connected components and holes. Holes represent the empty spaces in the data's shape, and connected components represent the connection of the data's points linked by a continuous path. Using the holes and connected components, we can calculate various topological properties to quantify a shape. For this, we perform the TDA on landslide shapes to compute topological properties which can then be used as a proxy to investigate the underlying failure {types}. It is important to note that we employ 3D point clouds (containing geographical latitude, longitude, and elevation information) from the landslide's outline (see Figures~\ref{fig:landsamples} and \ref{fig:datafig1}) obtained via the landslide polygon and the Digital Elevation Model (DEM). The landslide polygon provides the best available approximation of the landslide boundaries in the geographic space, as derived by standard surveying methods with suitable accuracy.

The degree of compactness in a landslide shape is essential when identifying failure {types} \cite{amato2021data}. For instance, slides are characterized by a more cohesive material that tends to remain as a single component (e.g., slides with clay-rich soil \cite{kenigsberg2020evolution}), leaving behind a more compact-shaped footprint as they fail. In contrast, flow type failures involve more fluid and fragmented materials deposited on a debris {fan} and display viscous/fluid kinematics that follows {the channelized topography} of the natural landscape and are less compact. On the other hand, falls consist of fragmented materials that roll/bounce off steep cliffs with a rather shorter and straighter run-out path compared to flows which leave behind a footprint that has an intermediate degree of compactness between the slide and flow-type failures. 

%We discovered that the average lifetime of holes—quantifying the empty space in the landslide shape—measures the degree of compactness. 
We use the amount of empty space inside the footprints of the landslide shape outline to quantify its compactness. To give a simple example, a higher amount of empty space in a landslide shape outline is associated with a higher degree of compactness (e.g., for slide-type failures as seen in Supplementary Figure S1-a). Representing the average lifetime of holes, AL\textsubscript{H} (one of the topological properties) computes the hole's average size and estimates the information pertaining to the empty space, and thus the compactness of a landslide's shape. So, landslide shapes with a longer AL\textsubscript{H} are more compact than shapes with a shorter AL\textsubscript{H}. Based on the Probability Density Function (PDF) of the AL\textsubscript{H}, our analysis reveals that slides are more compact than flows, falls, and complex landslides because they have a longer AL\textsubscript{H} (see Figure~\ref{fig:tdafeatures}). Moreover, we observe the AL\textsubscript{H} PDF curve for falls to lie between those of slides and flows, showing that empty spaces generated in falls do not survive long since materials detach from steep slopes and travel a short distance, thereby leaving behind a footprint that represents an intermediate level of compactness. Also, the PDF of the AL\textsubscript{H} for complex landslides shows an intermediate level of compactness, credited to their amalgamated behavior as a combination of slides, falls, and flows.

Another critical property for diagnosing failure types is the sinuosity of the transport zone, which describes the landslide's path or kinematic propagation as it progresses downslope. {Among various types of failure, flows exhibit the greatest sinuosity, following the channelized topography more closely than other mass-wasting mechanisms. This is attributed to the fluid and mobile characteristics of the materials involved.} In contrast, slides are the least sinuous as their material is rarely channelized and remains on the open slope, resulting in a relatively straight and uniform path. Fall-type failures are comparatively less sinuous than flows but still exhibit some degree of sinuosity, as they too follow the landscape's {topography}. Sinuosity defines the existence of numerous curves in the landslide’s shape attributed to the landscapes' {topography}, leading to the generation of partitions within the landslide outlines by the TDA and hence, generating multiple empty spaces with shortened lifetime (see Methods section). This information on the sinuosity of landslide shapes is inferred from the combination of two topological properties--the bottleneck amplitude of holes, BA\textsubscript{H}, and the average lifetime of holes, AL\textsubscript{H}. The BA\textsubscript{H} represents the maximum lifetime of holes in the landslide shape, which quantifies the maximum empty space in the 3D space occupied by the landslide. As sinuous shapes result in numerous smaller empty spaces with shorter lifetimes, the AL\textsubscript{H} drops without significantly impacting the largest empty space as determined by the BA\textsubscript{H} (see Figure~\ref{fig:tdafeatures}). Consequently, a landslide shape with a relatively higher BA\textsubscript{H} and a shorter AL\textsubscript{H} is indicative of increased sinuosity. In light of these observations, our findings indicate that flow-type landslides indeed display a higher degree of sinuosity compared to other failure {types}. This is evident from the fact that flows exhibit a similar BA\textsubscript{H} to falls but a shorter AL\textsubscript{H} in comparison. This is expected, as flows, being the most sinuous, cause multiple small holes or empty spaces (see flow-type and fall-type failures in Supplementary Fig. S1-b,c) that end up shortening the AL\textsubscript{H}. Conversely, slides display both longer BA\textsubscript{H} and AL\textsubscript{H}, reflecting their minimal level of sinuosity.

%Our findings indicate that flows and falls have similar PDFs of bottleneck amplitude of holes; however, flows have a lower lifetime of holes than falls. This behavior is evident as flows conjure smaller holes due to their higher sinuous kinematic path leading to a decrease in the average lifetime of holes. In contrast, slides have a higher bottleneck amplitude of holes and a higher average lifetime of holes, reflecting their low degree of sinuosity. 

We are also interested in the role that slope variations play, as they significantly impact the stability of the slope and influence the type of landslide that occurs. For example, falls and slides have a more significant slope transition in their profiles compared to flows, which propagate with a nearly constant slope \cite{catani2005landslide}. This slope variation is captured by the lifetime of the connected components. A sharp change in slope causes the points outlining the landslide to be spaced vertically further apart, leading to a longer lifetime of the connected components. Two topological properties—the Wasserstein amplitude of the connected components, WA\textsubscript{C},  and the average lifetime of the connected components, AL\textsubscript{C}—help capture information about this slope variation in a landslide's profile. The WA\textsubscript{C} quantifies the set of longer lifetimes of the connected components, quantifying the most significant slope change in the landslide outline. This is nicely illustrated in the PDF (Figure~\ref{fig:tdafeatures}) of WA\textsubscript{C}, which shows that slide and fall failures underwent more drastic slope changes compared to flows. Yet, falls possess a shorter AL\textsubscript{C} than slides. This is due to the lower portion of the shape's outline (at the talus) displaying a flatter terrain (representing the area where materials accumulate) and attributing negligible slope change, which ultimately shortens the AL\textsubscript{C}. In contrast, flows display the minimum AL\textsubscript{C}, as they more or less propagate on constant slopes.

\begin{figure*}[t!]
\includegraphics[width=0.9\columnwidth]{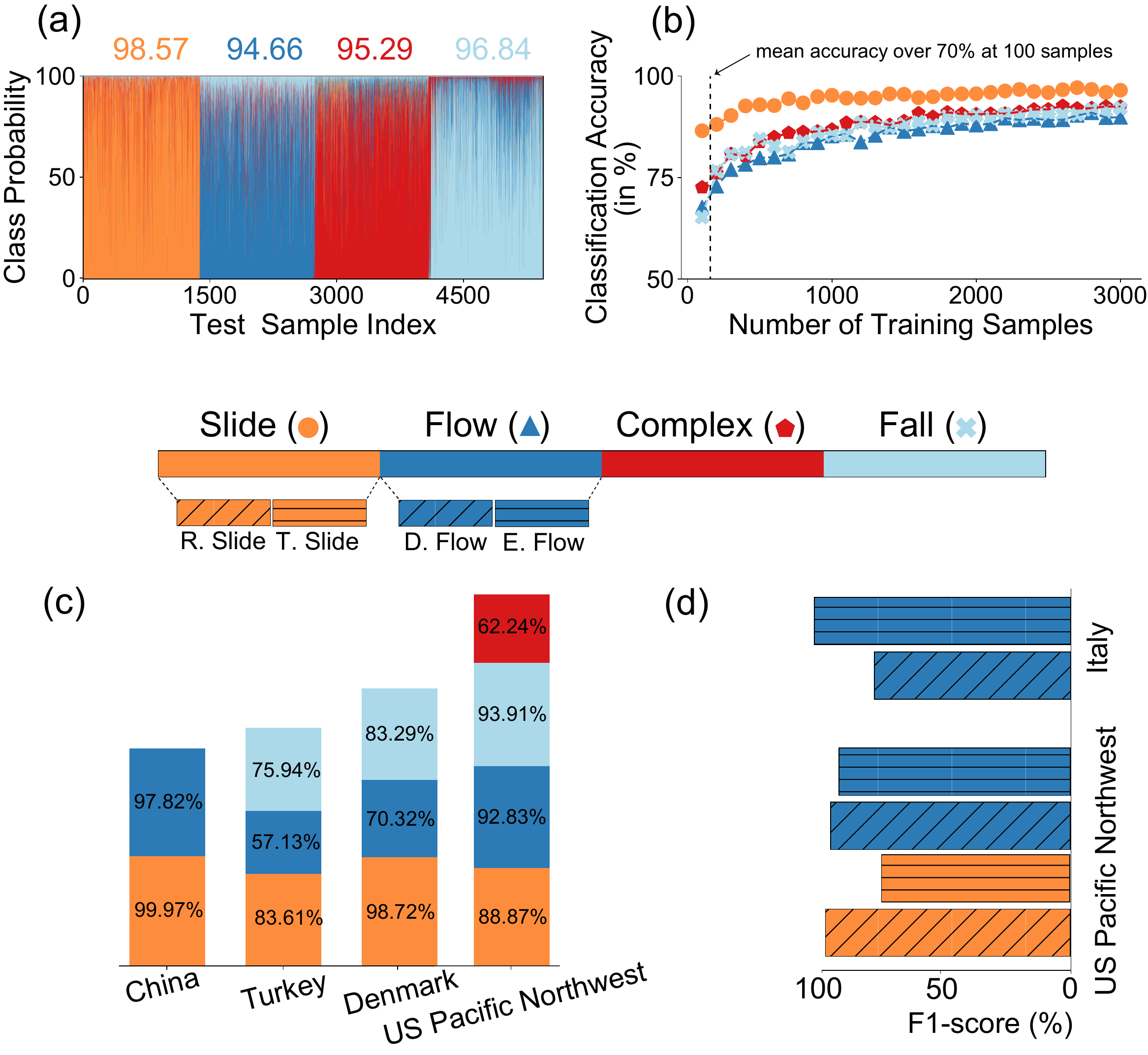}\hfill
\caption{Plot (a) shows the classification accuracy (in \%) for each failure class in Italy. The x-axis of the plots shows the testing sample's index, and the y-axis shows the class probability corresponding to each failure class. Plot (b) shows the classification accuracy (in \%) corresponding to each failure {types} with the number of training samples. The x-axis shows the number of training samples from each class used to train the model, and the y-axis shows the classification accuracy (in \%) corresponding to each class. At {100} samples, the mean classification accuracy {already} reaches over 70\% in Italy. {Plot (c) demonstrates the model's versatility in accurately identifying various types of landslides across multiple geographical regions, including China, Turkey, Denmark, and the US Pacific Northwest, with their corresponding F1-scores listed for each type. Plot (d) highlights the model's capability to distinguish sub-types of landslide failures specifically in the US Pacific Northwest and Italy. In the US Pacific Northwest, the model successfully classifies four additional sub-types—rotational slide, translational slide, debris flow, and earth flow—with an average F1-score of 84\% along with the other failure type classes (complex and fall type). In Italy, the model identifies two additional sub-types—debris flows and earth flows—with an average F1-score of 96\% along with the other failure type classes (slide, complex, and fall type).}}
\label{fig:result}
\end{figure*}

%It's worth noting that the same dataset, along with an equivalent number of training samples for other failure classes such as falls and complex failures, was used for both the US Pacific Northwest and Italy, while only slide and flow samples were further subdivided into their respective sub-types.

Several topological properties, like the Betti curve-based feature (BC), capture more intricate landslide shape properties and help in discerning landslide failure {types}. The Betti curve-based feature represents the total number, lifetime, and presence of the structures (holes and connected components) emerging simultaneously. We hypothesize that it encompasses a combination of compactness, sinuosity, slope variations, and similar structures within a given landslide shape. However, the exact connection to the underlying physical {types} is not clear due to the complex nature of this topological property. We anticipate that such properties consider higher-order information about the landslide shape that is not immediately apparent to us. 

Through our analysis, we discovered that common topological properties such as AL\textsubscript{H}, AL\textsubscript{C}, and BA\textsubscript{C} govern the general movements of distinct failure types. These properties act as proxies for the diverse kinematic and mechanical characteristics, which are essential to consider when identifying the different failure {types}. This finding simplifies and brings coherence to our understanding of landslide behaviors from a topological lens, offering a more effective approach to discerning and predicting these complex natural phenomena.

\subsection{Advantages of landslide topology over landslide geometry}

Traditional geometric descriptors of landslide shape, including properties such as area, perimeter, and convexity, are derived from 2D representations of the landslide body. As a consequence of this inherent simplification, these 2D-based geometric properties may not adequately capture crucial information, such as failure depth or internal deformations, associated with the landslides’ 3D configuration. To address these limitations and provide a more comprehensive understanding of the landslide dynamics, we computed topological properties that are derived from the landslides’ 3D configurations. We postulated that the topological properties would prove more meaningful in decoding the characteristics of the landslides and their underlying failure {types} than the traditional geometric counterparts. To test this, we used a set of seven well-known geometric properties that are commonly employed in the literature \cite{rana2021landslide, taylor2018landslide, stark2009landslide} along with six topological properties (for the justification of using six topological properties, please refer the Methods section) to determine the failure {types} in the Italian context. 

We jointly computed the feature importance of geometrical and topological properties using the Gini-index feature importance method in the random forest algorithm (see Figure~\ref{fig:tdafeatures}-g). After running over 100 iterations on the Italian data set, our findings consistently demonstrated that topological properties exhibited higher feature importance than the traditional geometric counterparts (achieving Micro F1-scores of 94\%, $\sim$65\% respectively), yielding superior predictive capabilities for identifying failure {types}. Additionally, we observed that even the least important topological property (BC\textsubscript{C}) has similar feature importance as the other geometric ones, while the former conveys unique information about the landslide shapes as discussed in the previous section. Moreover, we calculated the Probability Density Function (PDF) for both geometric and topological properties and observed that the latter had greater dissimilarity than the former among different failure types (see Supplementary Figure S2). These two findings demonstrate that topological properties are stronger predictors for identifying failure {types}. The reason for this can be attributed to the enhanced capacity of topological properties to encapsulate important information pertaining to landslide kinematic progression, failure depth, sinuosity, compactness, and slope variation.

% Landslide geometric properties, such as area and perimeter, based on 2D polygons, capture the landslide's physical dimensions. However, it misses important information to distinguish between distinct failure mechanisms, such as failure depth and internal deformations. On the other hand, landslide topological properties based on landslide 3D shape capture the complex morphology of a landslide, such as depth of failure, the width of kinematic propagation, and sinuosity making them a perfect choice to determine failure mechanism (for further details see Supplementary Section S3). Furthermore, using feature importance analysis and probability density functions (PDFs) for geometric and topological properties (see the supplementary information in section S2), we found that topological properties are excellent predictors for identifying failure mechanisms compared to geometric properties. For example, in Figure S2 (also see Figure~\ref{fig:tdafeatures}), the PDFs of topological properties like the average lifetime, bottleneck amplitude, and Wasserstein amplitude of holes showed more dissimilarity among different failure types than geometrical properties. %The PDFs were unique enough to detect substantial differences between each failure type when the topological properties/features are used.

\subsection{Determining failure types with TDA and machine learning}

Next, we employed Topological Data Analysis (TDA) to compute a diverse array of topological properties/features using the landslide inventory of Italy. Subsequently, we conducted a correlation analysis and feature importance assessment to identify the six most optimal properties out of thirty. Our evaluation unveiled that several TDA properties were redundant, amplifying the model's complexity and undermining its predictive potential. Consequently, we opted to eliminate the irrelevant ones. Leveraging the six best features, we applied the random forest algorithm to discern the failure {types}. We independently scrutinized the performance of our approach using various training and testing sets for the Italian inventory. 

We applied the model to approximately 250,000 landslide samples, ensuring balanced training by using an equal number of samples—13,000 for each landslide type. To mitigate overfitting and bias, we performed 10-fold cross-validation 1,000 times on a subset of 54,440 samples. This approach yielded a Micro F1-score, a key performance metric, surpassing 94\% for each failure type (see Figure~\ref{fig:result}-a), with a performance standard deviation of less than 0.2\%. This illustrated the robustness of our methodology in handling variations among training samples across Italy. We examined various other metrics (Supplementary Figure S3), such as the True Positive Rate (TPR) and True Negative Rate (TNR), to evaluate the method's performance. These metrics consistently exhibited high scores across all classes, thereby ascertaining the model's classification ability.

\subsection{Method transferability to different regions of the world}
%To evaluate the method's effectiveness and applicability in real-world scenarios, we tested the proposed approach's efficiency in various geomorphological, geological, and climatic settings. In addition, we tested the approach on the event-specific landslide databases by utilizing spatiotemporal inventory from Wenchuan, china. Furthermore, we illustrate the approach's useability in real-world scenarios when a region of interest lacks adequate or has no landslide samples with failure-type information for training the method. 
{In this sub-section, we assess the performance of our method's transferability across both historical and event-based landslide inventories that include spatial and temporal data. Additionally, we tested the method's efficacy in scenarios where information on failure types is either scarce or lacking.}
{To evaluate the method transferability in diverse geomorphologies and climatic regions, we deploy the method in the Pacific Northwest of the US (states of Oregon and Washington), Denmark (national inventory mapped by Luetzenburg et al.\cite{luetzenburg2022national}), Wenchuan (China) inventory mapped by Fan et al.\cite{fan2019two} for the 2008 earthquake event, and Turkey (municipality of Ulus) mapped by Gorum et al.\cite{gorum2019landslide}.} {The method achieved above 80\% mean classification accuracy in each region in identifying failure types (see Figure~\ref{fig:result}-c). 
}

\begin{figure*}[t!]
\includegraphics[width=\columnwidth]{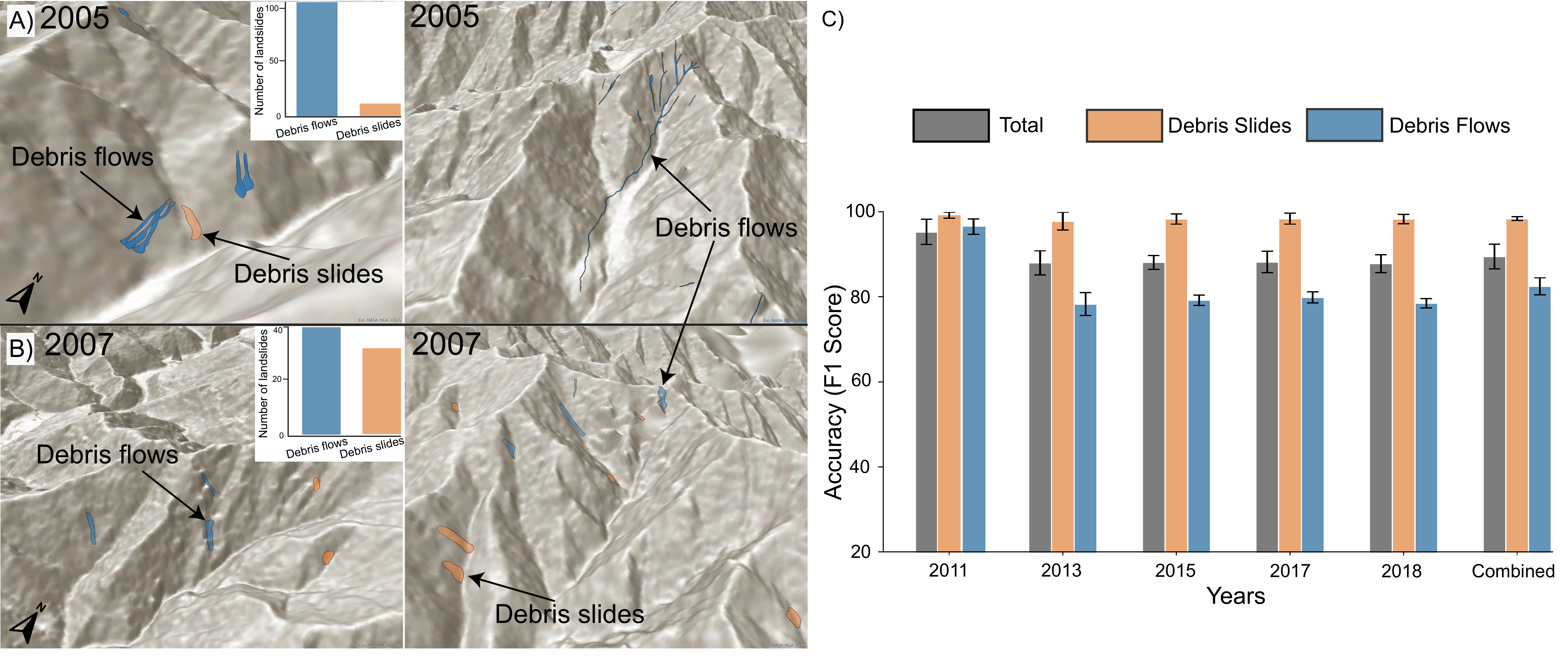}\hfill
\caption{{Rows (a) and (b) display snippets of proposed method predictions in two undocumented databases from the $2005$ and $2007$ inventories of Wenchuan, China, respectively. Each row in these plots shows debris slides and flows as identified by the model, with the accompanying bar chart quantifying the number of landslides by their type. Plot (c) shows the method performance on multi-temporal databases triggered by either earthquake or rainfall spanning from 2011 to 2018 years. The model is trained on the 2008 inventory associated with that year's co-seismic event for each testing year. } {The chart also includes a composite dataset derived from combining all these multi-temporal inventories. Error bars in the chart indicate the standard deviation for each landslide type within each inventory. The model achieved an average F1-score of 89\%, with an average standard deviation (for the F1-score) of 4\%. The base map was sourced from the World Hillshade Map. Map credits: World Shaded Relief-ESRI\cite{ESRI2020}.}}
\label{fig:temporal}
\end{figure*}

{Moreover,} { we implemented the method in an event-specific inventories from the multi-temporal inventory from Wenchuan, China containing eight event-specific inventories from distinct years, triggered by a combination of rainfall and earthquakes. To assess the method's performance on an event-specific inventory, we trained the model using one of the earthquake-triggered inventories from 2008 and then tested it on the remaining inventories from the subsequent years (2011 to 2018). Across all the testing inventories, our method achieved a mean classification accuracy of over 90\% (see Figure~\ref{fig:temporal}-c).} {Additionally, we implemented the method on two undocumented inventories predating the 2008 event, specifically from the years 2005 and 2007. In our prediction of the landslide types within these inventories, we found that most were identified as debris flows (119 in 2005 and 40 in 2007), with the remaining few classified as debris slides (13 in 2005 and 32 in 2007). We manually verified the method's predictions for these landslides using ArcGIS and Google Earth archive imagery (see Figure~\ref{fig:temporal}-a, b).} 

% {Overall, the consistency of our method underscores the model's versatility in handling both historical and event-based inventories and emphasizes its potential for updating undocumented historical landslide records and rapidly identifying the failure-types of new landslides triggered by recent events, respectively.}

{To assess our method's performance in regions with limited documented failure-type samples, we conducted an experiment with a decreasing number of training samples. We discovered that with just 100 samples from each class, our method achieved a mean classification accuracy exceeding $70$\% {using the Italian dataset} (see Figure~\ref{fig:result}-b). This highlights our approach's capability to operate effectively in regions where failure-type samples are scarce. For areas entirely lacking documented failure information, the model can be used by manually annotating a modest set of around 100 samples. This provides a practical solution for a streamlined, automated failure type classification system for documenting failure types in large landslide databases, often containing thousands of entries.}

\subsection{Determining sub-type failure movements of the landslides}
{
Sub-type classification of landslides is critical for devising targeted mitigation strategies and understanding the underlying mechanics that drive these failures. To address this need, we go beyond the general categorization of landslide failure types to test the ability of the model to identify/classify specific landslide typologies for the inventories of Italy and the Pacific Northwest of the USA. In the case of Italy, we tested with five sub-type failure movements ('Slides', 'Debris flows', 'Earth flows', 'Complex landslides', and 'Rock falls') achieving an F1 accuracy of 96\% while in the case of the Pacific Northwest of the US, we tested with six sub-type failures ('Debris flows', 'Earth flows', 'Rotational slides', 'Translational slides, 'Rock falls', and 'Complex landslides') achieving an average F1 accuracy of 84\% (see Figure~\ref{fig:result}-d). This extension demonstrates that our method can handle more than the basic four classes or types, including differentiating sub-types within slides (such as rotational and translational) and flows (including earth flow and debris flow). Such distinctions are vital for type-specific predictive modeling and a comprehensive understanding of landslide movements, as well as the underlying causes of their behaviors. This success can be attributed to the TDA's capability of discriminating among various failure types like debris flow and earth flows. This differentiation arises from the inherent morphological variations between debris (characterized by more sinuous and longer trails) and earth flows (marked by less sinuous and relatively shorter trails).}

\subsection{Identifying failure types within the complex landslides}

Complex landslides typically occur as amalgamations of numerous processes or failure types that appear successively, such as slides to flows \cite{lahusen2020rainfall}. Because complex landslides include multiple failures, it is challenging to investigate their behavior for the purposes of predictive modeling. Topological properties are capable of capturing intricate information between different failure processes (as seen in the previous sections), and hence, we further explored its capability to understand the underlying coupled failure types that form complex landslides.
%to investigate the combination of mixed failure mechanisms that form complex failures. %We wanted to explore the power of our approach to interpreting the existing landslide failures that are marked as "complex" in the US Pacific Northwest.
We utilize 428 complex landslides from the US Pacific Northwest data set to discern the combination of failure types present in them. Out of 428 complex landslides, 198 of them are documented as "Translational rock slides followed by rock falls" and the rest are documented as "Rotational slides followed by flows" (as reported by the Statewide Landslide Information Database for Oregon, SLIDO \cite{franczyk2019statewide}).

To identify the failure {types} within these complex failures, we trained our method with three classes (i.e., slides, flows, and falls) and forced the model to predict the class probability corresponding to each failure type. For 198 complex landslides documented as "Translational rock slides followed by rock falls", our model predicts slide-type failures with the highest probability followed by falls (see Supplementary Figure S4-d). Similarly, for the remaining 230 "Rotational slides followed by flows" complex landslides, our model predicts slide-type failures with the highest probability followed by flows (see Supplementary Figure S4-e). Among these 428 landslides, sliding {failures} are predicted as the most dominant failure type, which is also evident when observing the resemblance of the slide and complex failure topological properties (such as AL\textsubscript{H}, BC\textsubscript{H}, and BA\textsubscript{H}) in the PDF plots (see Supplementary Figure S4-a,b,c). These findings demonstrate that topological properties are able to capture more than just one physical process in a given landslide and that they could be used to automatically improve large Varnes-based inventories toward Cruden and Varnes-based classification.

\section{Discussions}
In this work, we attempt to determine the failure types based on their movements through the lens of landslide topology. Our key findings elucidate the connection between the topological properties as a proxy to identify underlying failure movements. We observe that identical landslide types harbor similar topological properties, indicating the presence of common morphological characteristics that govern the general movement of the failures. {We find that topological properties offer a more profound capacity to distinguish between failure types than traditional geometric properties {(please see result sub-section on landslide topology as a proxy to identify failure movements)}. This finding can be attributed to the fact that topological properties inherently capture critical information related to landslide kinematic progression, failure depth, sinuosity, compactness, and variations in slope. In contrast, geometric properties tend to oversimplify the complex spatial, kinematic, and mechanical relationships that govern the behavior of landslides and are hence less effective in differentiating between various failure {types}. Building on the advantages of topology over geometry, we developed a method using the Italian landslide inventory that utilizes topology to identify failure types.}

{The next valuable step involved recognizing our method's capability to adapt across diverse geomorphological and climatic settings, including both historical (such as Italy, the US Pacific Northwest, Denmark, and Turkey) and event-based inventories (such as Wenchuan, China). This exploration yielded noteworthy results, as our method is capable of identifying failure types across these varied regions. This opens up opportunities for transferability and provides a path toward identifying failure movements in previously undocumented inventories. Consequently, recent events and their corresponding inventories would benefit from this approach, as they could be efficiently classified by employing our method.}

{Beyond providing a transferable method, our approach also opens new possibilities for studying the complexity and evolution of landslide behavior, with the potential to improve hazard forecasting \cite{rana2021landslide}. As demonstrated in the temporal experiment of Wenchuan, China, our approach admittedly holds the ability to identify failure movements across not just space but also time, with which we were able to also recognize and quantify the failure movements in two undocumented inventories predating the Wenchuan earthquake event (Mw 7.9) of 2008 (please see the transferability results sub-section). Such information carries substantial importance as it paves the way for focused research into quantifying (re-)mobilization of failures, extracting precise data on sediment budgets, and understanding dominant geophysical cycles at continental and global levels \cite{fan2019earthquake, mirus2017hydrologic, pearce1986effects,east2020geomorphic}.}

{Drawing on the experiments of the sub-type classification {(for e.g., debris and earth flows)} of landslide failures {using the Italian and the US Pacific Northwest dataset (please see the sub-type results sub-section)}, we recognize the substantial value in identifying and quantifying these sub-movements, bearing notable potential to enhance both landslide risk assessment and related hazard models \cite{huang2023uncertainties}. The level of damage to infrastructures and the risk of human casualties vary depending on the intensity of the failure movement, which differs for each failure type \cite{varnes1978slope}. For example, a slow-moving deep-seated rotational landslide (1.5 m/year to 16 mm/year) may not pose an immediate threat to the population, but it can cause extensive structural damage to buildings over a prolonged period \cite{sundriyal2023brief, dille2022acceleration}. In contrast, flow-type failures, such as debris flows, have rapid mobility and can result in significant casualties and infrastructure damage simultaneously \cite{vega2016quantitative, perkins2012death}. Similarly, episodic impacts in fall-type failures can cause massive damage to infrastructures in a matter of seconds due to their high energy (e.g., impact pressure measured in kilopascals, kPa) \cite{dietze2017seismic}. We can infer from these broad examples that the availability of failure-type, especially sub-type information could considerably enhance the accuracy of predictive modeling and that incorporating it benefits the landslide community as it enables the development of accurate landslide predictive models.}

{In addition, we utilized topological properties to dive deeper into complex landslides and identify the underlying coupling of failure types that contribute to their formation {(please see "Identifying failure types within the complex landslides" Results sub-section)}. Our findings in the US Pacific Northwest suggest that topological properties can reveal more than one physical process in a given landscape (for example, identifying coupled failures of slides following falls or slides following flows). This has significant implications for understanding complex landslide failures, which often arise from a combination of different failure movements, such as sliding, flowing, and falling, and may not be fully encompassed by current characterization and classification methods for large-scale analysis. Traditional approaches may struggle to pinpoint the exact cause, leading to hindrances in prevention efforts. The use of topological properties to uncover these intricacies offers a path towards a more comprehensive characterization, where capturing the dominant failure type would enable pinpointing the possible initial failure within the complex landslide. Given these advantages, we anticipate that our method will open new avenues for future research, particularly in the landslide modeling community, for example, working on large-scale post-event mapping and large- to medium-scale hazard forecasting.}

While the proposed solution demonstrates notable success in identifying failure {types}, there are inherent limitations that warrant attention. The effectiveness of the method relies on the quality and geographical location of the training and testing data used in the model. Manual annotations of failure types can lead to bias since various mappers will have different perspectives (mapping on aerial or satellite imagery versus geomorphological field mapping can display distinct perceptions) when annotating the landslides and their failure types. Also, due to the ambiguous nature of complex-type failures, they could include slides and flows simultaneously, which can impact the overall performance of the model. The method's reliance on a DEM for converting 2D polygons into 3D shapes also presents potential challenges. DEM quality in the training and testing regions can bias the results, particularly for smaller landslides, as coarser DEM resolutions may struggle to capture the profiles of these smaller-scale events.

{The potential of the proposed method reaches beyond just understanding the complex interplay between landforms, their shapes, and the underlying geophysical processes responsible for their formation; they also serve as a subject captivating interest across various geophysical disciplines. The ability to acquire knowledge about the processes generating complex landforms based solely on their shapes suggests a rich presence of signatures imprinted on the landscapes. Our method leverages their topological properties to effectively extract this information. Envisioning compelling applications beyond landslides, we can explore other geophysical processes such as permafrost-borne retrogressive thaw slumps in Arctic regions \cite{nicu2021preliminary} and sub-surface processes, e.g., submarine landslides \cite{frey2006frontally}, which commonly occur in a typical data-scarce environment such as the sea bottom, where geological and geotechnical information are almost absent. These processes also give rise to unique landforms displaying distinct shapes and configurations, and employing topology can aid in gauging the mechanisms governing their occurrences. By doing so, our method could shed new light on mathematical and physical phenomena underlying various geophysical and environmental scenarios.}

\section{Data availability}\label{sec6}
The dataset we utilized in this study to classify the failure {types} of landslides was obtained from the Inventario dei Fenomeni Franosi (Inventory of Landslide Phenomena) in Italy (IFFI) \cite{trigila2010quality}. The IFFI project catalog (\url{www.progettoiffi.isprambiente.it}) was created in $1999$, with the aim of mapping and identifying landslides in Italy, and holds information on over $250,000$ usable landslide polygons. 
Aerial image interpretation, historical sources, and field surveys were used to acquire and validate this catalog, while the classification protocol/scheme referred to that of Varnes ($1978$) \cite{varnes1978slope} and Cruden and Varnes ($1996$) \cite{cruden1993cruden}. In our work, we chose the polygonal landslide data from this catalog and also carried out post-processing to correspond to the spatial extent and resolution of the $25$-meter EU-DEM \cite{bashfield2011continent}.

The study area is the Pacific Northwest region of the United States. The data set from the US Pacific Northwest consists of inventories from Oregon's Statewide Landslide Information Database for Oregon (SLIDO-4.4 updated $10/29/2021$; Franczyk et al.\cite{franczyk2019statewide}), mapped by the Oregon Department of Geology and Mineral Industries (DOGAMI), and the Washington State Landslide Inventory Database (WASLID updated $2018/08/01$; Slaughter et al.\cite{slaughter2017protocol}), mapped by the Department of Natural Resources, Washington Geological Survey (WGS). The combined inventories comprise $47,653$ landslides from the US Pacific Northwest region. The inventories contain LiDAR-derived landslide polygons guided by protocol to capture the movement types with spatial information on the scarps, head scarps, toes, and deposits \cite{burns2009protocol,burns2016protocol,slaughter2017protocol}. Since this data is categorized using a combination of Cruden and Varnes \cite{cruden1993cruden} and Hungr et al. \cite{hungr2014varnes} (i.e., slides, flows, complex, and falls), we modified the Italian data correspondingly to maintain uniformity in the taxonomy of the failure mechanisms.

{The national landslide inventory of Denmark was obtained from Luetzenburg et al.\cite{luetzenburg2022national}, published in 2022. The landslides were mapped via high-resolution DEM of 2015 and orthophotos supplied by the Danish Agency for Data Supply and Infrastructure, consisting of 3202 unique polygons of mapped landslides following the classification from Hungr et al.\cite{hungr2014varnes}.} 

{The landslide inventory of Turkey was obtained from Gorum\cite{gorum2019landslide} published in 2019. The landslides were mapped from airborne LiDAR data with a total count of 900 landslides classified according to the Cruden and Varnes\cite{cruden1996landslides} system.}

{Landslide inventory of the Wenchuan region of China was acquired from Fan et al.\cite{fan2019two} where they generated a multi-temporal inventory of the infamous Wenchuan 2008 earthquake event causing thousands of landslides. The multi-temporal window spans from 2005 to 2018, mapped with a custom classification system based upon a simplification of the Hungr et al.\cite{hungr2014varnes}.} 
The EU-DEM for Italy and {Denmark} was downloaded from \url{https://land.copernicus.eu/imagery-in-situ/eu-dem/eu-dem-v1.1} and the DEM for the US Pacific Northwest was downloaded from \url{https://www.opentopography.org/}. {The Shuttle Radar Topography Mission (SRTM) DEM for China and Turkey was downloaded from \url{https://dwtkns.com/srtm30m/}}
% We converted the landslide database shapefiles and the DEM into the WGS84 geographic coordinate system for data and feature engineering purposes.

\section*{Code availability}\label{sec7}
\sloppy

Data analysis and processing were conducted using Python programming language and its associated libraries. The various scripts used for data analysis are available at \url{https://github.com/kushanavbhuyan/Uncovering-landslide-failure-types}.

\section*{Author Contributions}\label{sec8}
KR, KB, UO, and NM contributed to the conceptualization and design of the research. JVF and KB curated the data. KR and KB developed the methodology and conducted the formal analysis under the close supervision of UO, FCa, FCo, and NM. All authors contributed to writing, review and editing.

\section*{Acknowledgments}\label{sec8}
We extend our gratitude to Dr. Tolga Gorum and Dr. Hakan Tanyas for providing access to the Türkiye landslide inventory dataset, which served as an essential resource for empirically evaluating the model’s transferability.

\section*{Competing interests}\label{sec10}
The authors declare no competing interests.

%\bibliographystyle{unsrt}  
%\bibliography{references}  

\clearpage
\begin{center}
\textbf{\large Supplementary Materials: Landslide Topology Uncovers Failure Movements}
\end{center}

\setcounter{section}{0} % Reset the section counter
\renewcommand{\thesection}{S\arabic{section}}

\setcounter{figure}{0} % Reset the figure counter
\renewcommand{\thefigure}{S\arabic{figure}}

\section{Introduction}
This supporting information (SI) to the manuscript titled “Landslide topology uncovers failure movements”  includes a detailed analysis of landslide topology and its importance in finding the landslide failure {types} based on their movement. The SI includes an in-depth analysis of the topological features and their probability distributions, quantifying the differences among the different failure {types} and the connection between the landslide topology and the physical processes. Moreover, we include a section for a detailed evaluation of the model and its transferability analysis. We also show that landslide topological information provides more information about landslide shape than classical geometric information like area, perimeter, and ellipticity, and therefore can be helpful in other landslide research. 

\section{Behavior of different failure {types}}

\begin{figure*}[t!]
\includegraphics[width=\columnwidth]{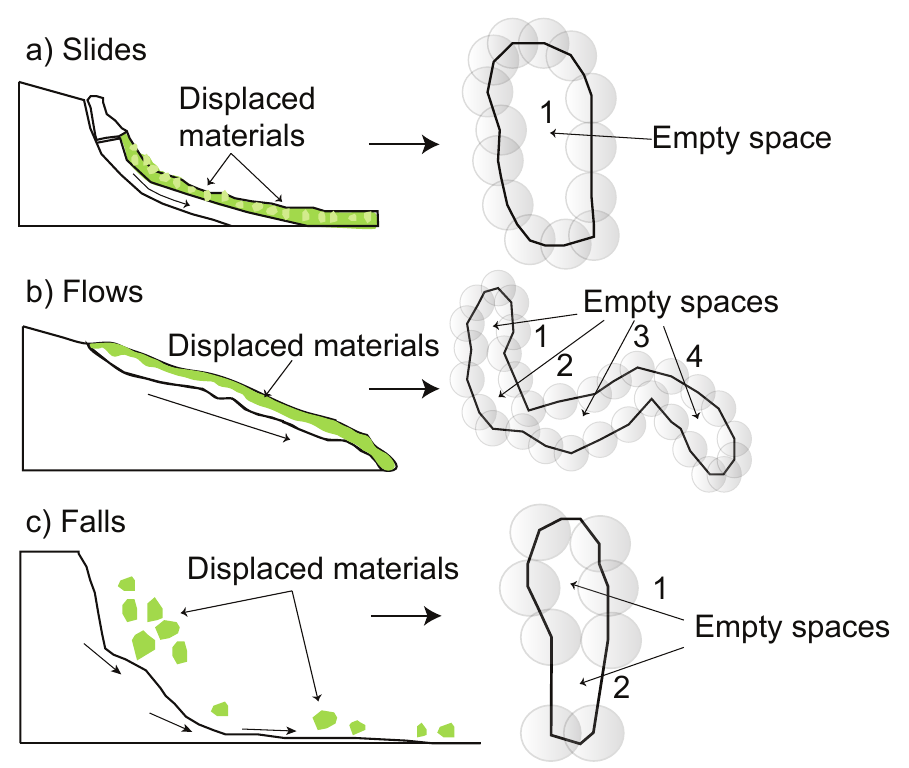}\hfill
\caption{The diagram shows a schematic outlook of how empty spaces are created in different landslide types. This illustration is shown using a simplified 2D transformation of a rather complex 3D topological phenomenon for ease of understanding. Sub-plots (a–c) refer to the possible configurations of empty spaces created in the typical polygons of each failure {type}. Slides tend to have the fewest empty spaces or holes due to their compact shapes, followed by falls. Flow-type failures tend to have multiple numbers of empty spaces due to the sinuous shapes they conjure as they follow the landscape's channelized topography.}
\label{fig:concepTDA2D}
\end{figure*}

The inherent differences between failure {types}, notably their kinematic and mechanical behaviors, contribute prominent intricacies to the topography (see surface profiles in Figure 1 in the main manuscript). These intricacies are attributed to slope deformity, interior deformation, kinematic width of failures while propagating down-slope, main scarp deformation, run-out length represented by the debris/earth/soil transportation, and accumulated debris, and others captured by topology. The following are some of the most common failure {types} and their various behaviors.
%Many landslides involve several movement episodes separated by extensive or brief periods of relative inactivity. 

%Characterisitics of slides
The profile of rotational slides is marked by a conspicuous primary scarp and a distinctive back-tilted bench at the head, but little interior deformation (a schematic view can be seen in Figure 1 in the main manuscript). They are typically slowly moving a large portion of the weak rock mass. At the same time, kinematically rapid planar sliding is marked by the sliding of a rock mass on a planar rupture surface with little to no internal deformation, where the scarp might be separated from the stable rock at deep vertical tension cracks. Typically, they exhibit very compact shapes. Cohesion, $c$ plays an important role in slides, as the degree of internal strength between the particles in a block of material determines the strength and stability along the slip plane. Translational landslides, like the ancient Seimareh slide in Iran's Zagros Mountains, are among the largest and most destructive on Earth \cite{roberts2013gigantic}.

%Characterisitics of flows
Flows are characterized by very rapid movements consisting of saturated granular material on moderate slopes, including liquefaction of materials (in the context of co-seismic triggers) or excess pore pressure (in the context of rainfall triggers) originating from the landslide source. When the internal friction angle, $\varphi$, is low (due to the mixture of solid and fluid particles), less external force is required to instantiate a failure because they are displaced quite easily. Kinematically flow-ish movements are observed with channelized streams and a bulk deposit of debris at the talus (deposition zone), representing highly elliptical, elongated bodies \cite{evans2001landslides}.

%Characterisitics of falls
Usually limited in volume, falls (particularly rock falls) exhibit ballistic movements (high velocity, energy, and momentum) that are massively destructive. They detach from cliffs and move at high velocities, either by rolling, falling, or bouncing due to the influence of gravity. The run-out of a rock fall is often shorter and is more likely to travel along a straight path, whereas the run-out of debris flows is longer and can meander and spread out over a wider area \cite{bourrier2013use}.

%Characterisitics of complex failures
Complex failures are very hard to describe, as there is an amalgamation of different failure {types} occurring at the same time or subsequently, and they can therefore exhibit multiple characteristics of other {failure types}. For example, irregular debris slides evolving into a debris flow or any other combination of slides, flows, and falls eventually evolving into another movement style can be considered examples of complex failure \cite{cruden1993cruden,hungr2014varnes}. 

Such morphological and geometrical information for each distinct {failure type} is theorized to be captured in the topological space by the topological properties, which are then utilized in the machine-learning model to identify the failure {types}.

\section{Topological Features}
\begin{figure*}[t!]
\includegraphics[width=\columnwidth]{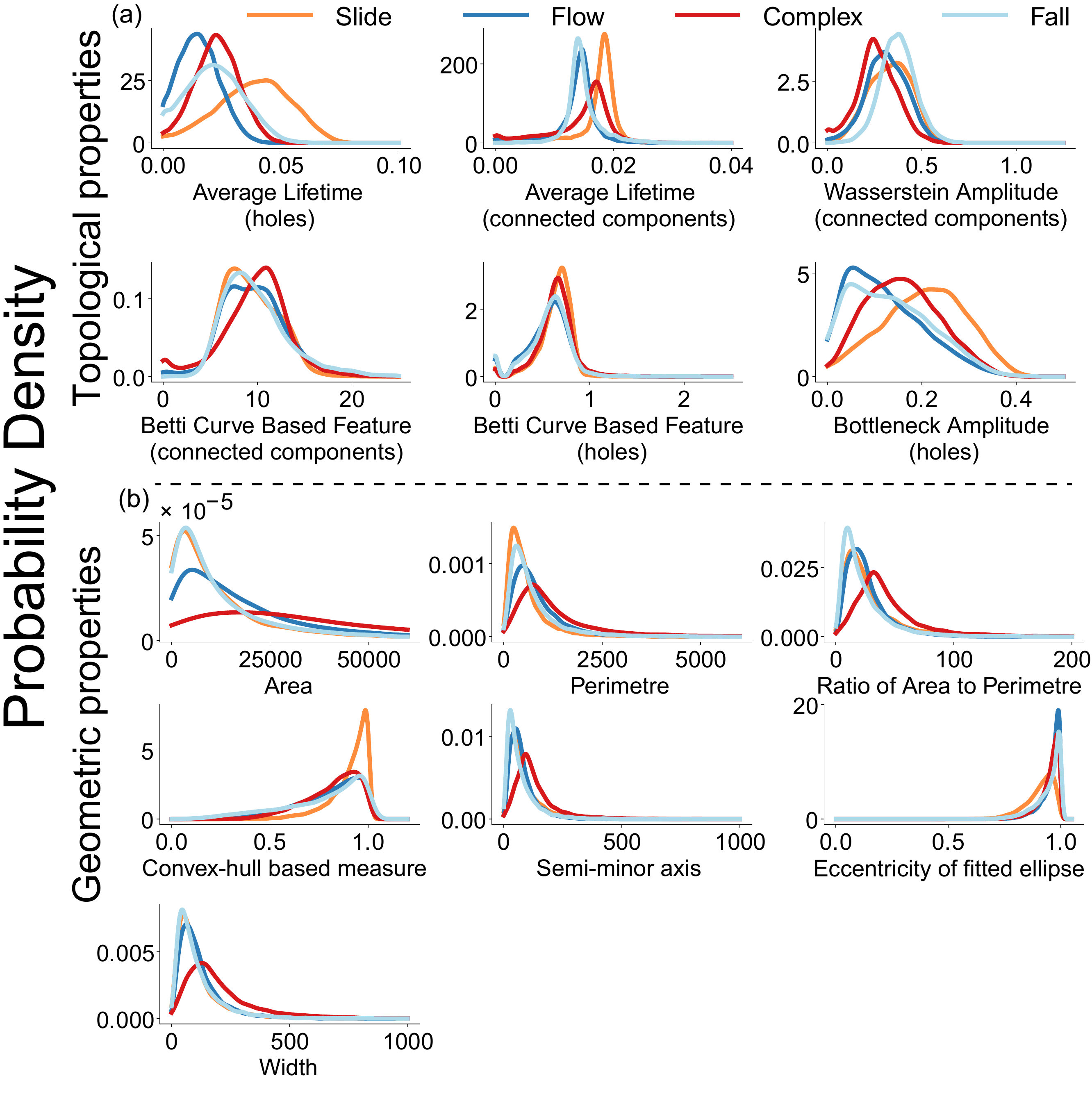}\hfill
\caption{Probability distribution functions of the geometrical and the topological features for each failure type-- slides, flows, complex, and falls--in Italy. The y-axis shows the probability density values (calculated using kernel density estimation), and the x-axis shows the value of topological or geometrical attributes. The geometric features from top to bottom are: area ($A$), perimeter ($P$), the ratio of area to perimeter $\frac{A}{P}$, convex hull-based measure ($C_h$), minor($s_m$),  and width ($W$) of the minimum area bounding box fitted to the polygon.}
\label{fig:geom_topo}
\end{figure*}

Persistence diagrams capture the life-death information of structures like connected components, holes, and voids. The persistence diagram consists of a set of $\{(b_i,d_i)\}_{i=1}^{i=N}$ pairs corresponding to each structure type; here, $i$ and $N$ are the indexes of birth-death pairs and the total number of the birth-death pairs. Using the set of $\{(b_i,d_i)\}_{i=1}^{i=N}$ pairs, we can calculate various topological features such as persistence entropy, average lifetime, number of points, Betti curve-based measure, persistence landscape curve-based measure, Wasserstein amplitude, Bottleneck amplitude, Heat kernel-based measure, and landscape image-based measure.

Some of the above topological features can be explained using a lifetime vector that is calculated using a set of $\{(b_i,d_i)\}_{i=1}^{i=N}$ pairs. The lifetime vector $[l_i]_{i=1}^{i=N}$ is calculated as the difference between death and life of the $(b_i, d_i)$ pair ($l_i=d_i-b_i$). The number of points, average lifetime, and persistence entropy are the length, average, and Shannon entropy of the lifetime vector. In comparison, topological features like Bottleneck and Wasserstein's amplitudes quantifying the magnitude of the lifetime vector are $p$-norm ($p$=2) and $\infty$-norm of the lifetime vector, respectively.

The Betti curve-based feature is a $p$-norm of a 1D discretized betti curve, which is a function ($B(\epsilon): {\mathbb{R}} \rightarrow \mathbb{Z}$) mapping the persistence diagram to an integer-valued curve, and it counts the number of birth-death pairs at a given $\epsilon$, satisfying the condition $b_i<\epsilon<d_i$ \cite{garin2019topological}. Similarly, a persistence landscape curve-based feature is a $p$-norm of a 1D discretized persistence landscape curve defined as $ \lambda(k,\epsilon): {\mathbb{R}} \rightarrow  {\mathbb{R}}_{+}$, where $ \lambda(k,\epsilon) = k_{max} \{f_{b_i,d_i}(\epsilon)\}_{i=1}^{i=n}$, $k_{max}$ is the $k$-th largest value of a set of functions defined by $f_{b_i,d_i}(\epsilon)$ = $max \{0,min(\epsilon-b_i,d_i-\epsilon)\}$ for each $(b_i,d_i)$ pair \cite{bubenik2017persistence}.

The heat kernel-based feature is a $p$-norm ($p$=2) of the discretized 2D function obtained using the operation of the heat kernel on the persistence diagram. Heat kernel uses a gaussian kernel ($\sigma$)  and a negative of the gaussian kernel ($\sigma$) for each $(b_i,d_i)$ pair and mirror of $(b_i,d_i)$ pair across the diagonal \cite{reininghaus2015stable}. In contrast, the persistence image-based feature is a p-norm (p=2) of the discretized 2D function obtained using the operation of the weighted Gaussian kernel on all $(b_i,d_i-b_i)$ pairs in the birth-persistence diagram \cite{adams2017persistence}. The birth-persistence diagram consists of $(b_i,d_i-b_i)$ pairs where the x-axis shows the birth information, and the y-axis shows the lifetime of the $(b_i,d_i)$ pair.

\section{Geometric versus Topological Features}
\begin{figure*}[t!]
\includegraphics[width=\columnwidth]{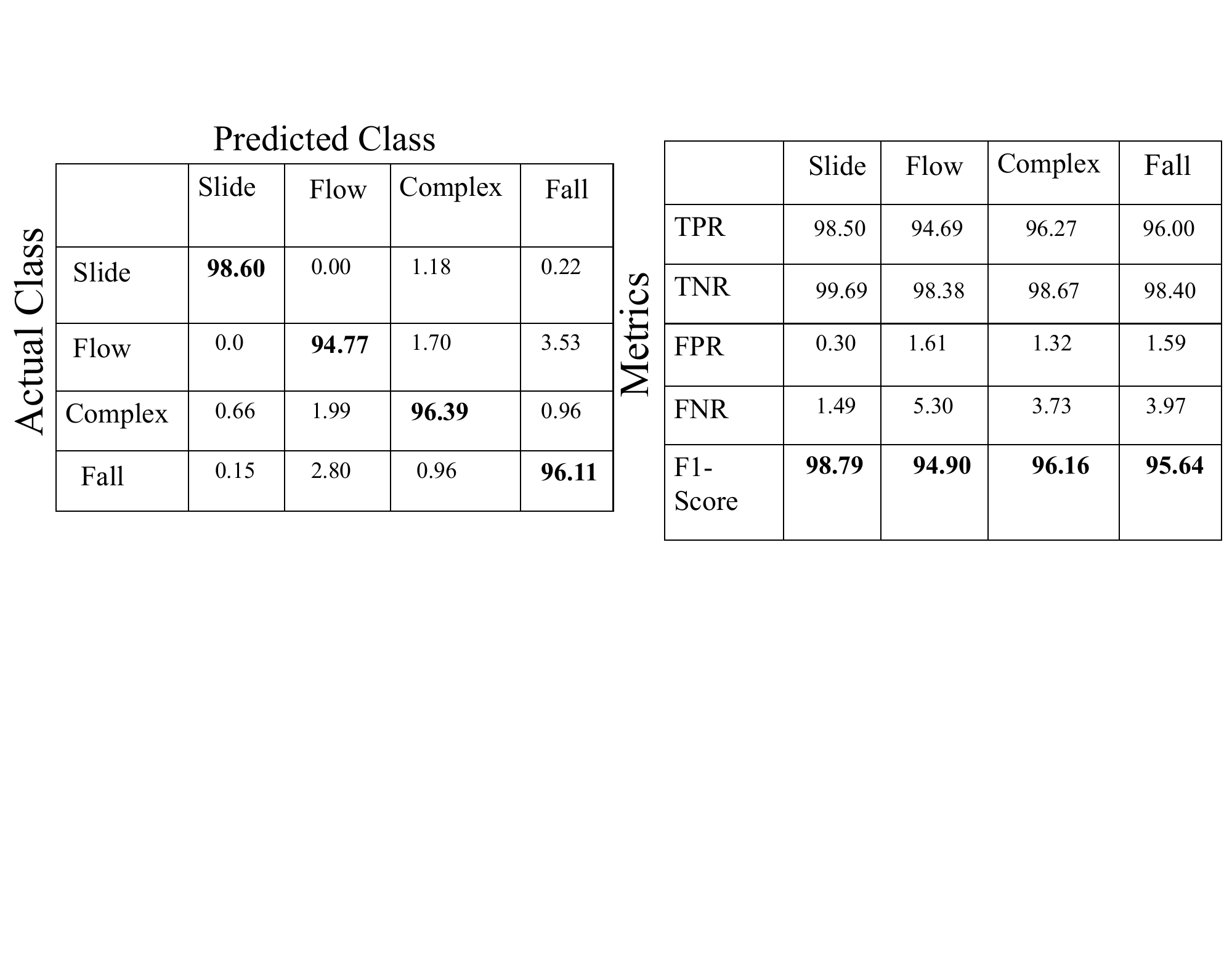}\hfill
\caption{The figure shows the confusion matrix and the associated accuracy metrics of the random forest model for the data of Italy.}
\label{fig:confusion}
\end{figure*}
Geometric properties define an object's shape and size, but topological properties explain the connections and topological interactions among its parts. Geometric properties such as area, perimeter, convexity, and ellipticity define the physical dimensions of a landslide, whereas topological properties such as the average lifetime of holes, Betti curve, and landscape curve describe the connections and interactions of the soil and rock masses, the width of kinematic propagation, and depth of failure in a landslide. Geometric properties are, however, sensitive to any changes made to the original shapes of the geometry and therefore, more susceptible to drastically changing the geometric property values. For example, any change to a landslide's boundary/body would inadvertently change each of the values of the geometric properties like area, perimeter, convexity, etc., but the same cannot be said for topology as it relies on the number of voids that are generated based on the overall shape of the landslide body. This is even more pronounced upon investigating the landslides in 3D. Since these geometric properties cannot be broadcasted to 3D, much information related to variational changes in the topography (attributed to elevation and slope) is lost. As TDA captures this 3D information and utilizes it when engineering topological features, intricate information on landslides such as depth of failure, deformations pattern, and the width of kinematic progression are well recorded.

\begin{figure*}[t!]
\includegraphics[width=\columnwidth]{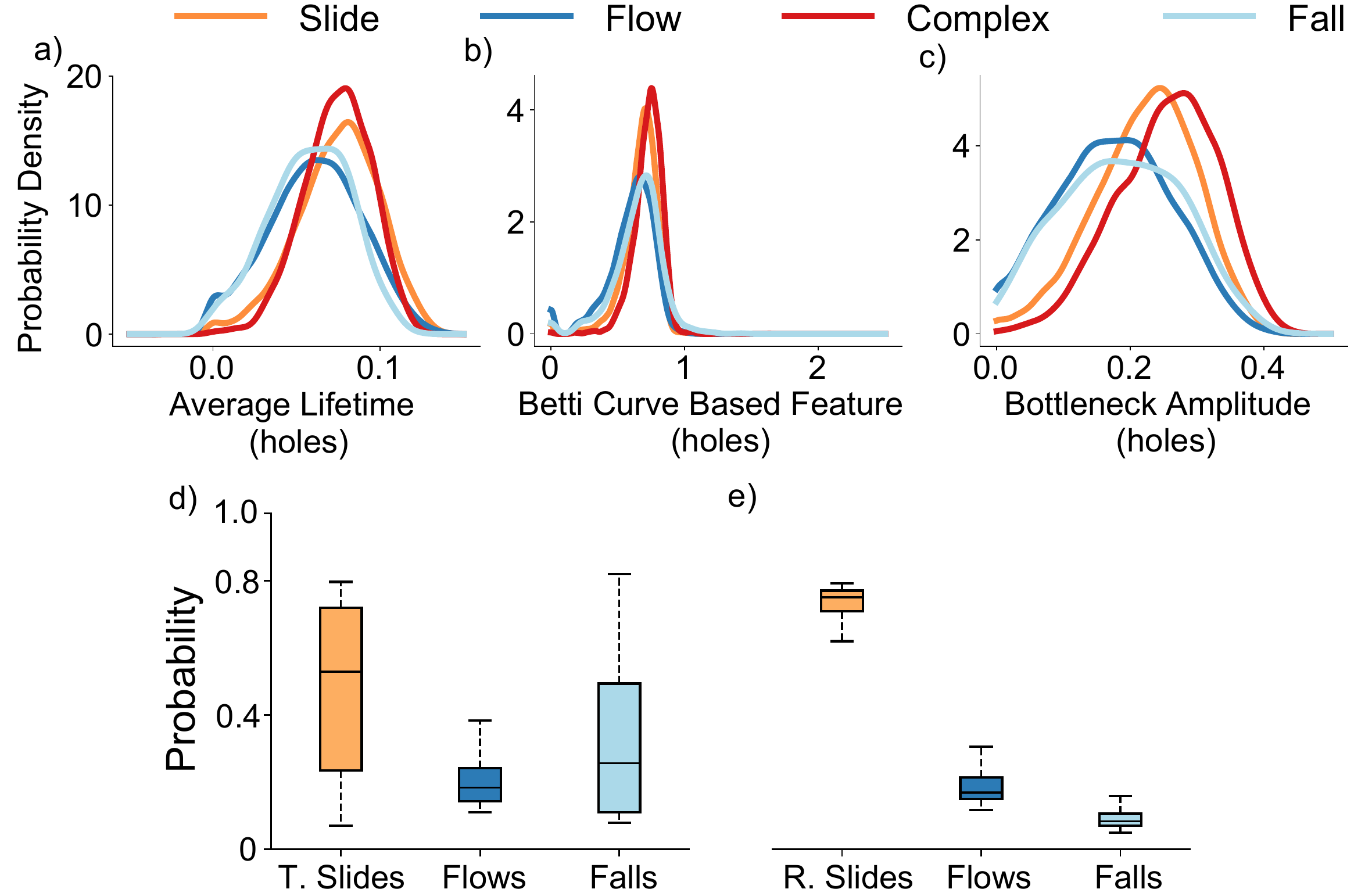}\hfill
\caption{{Sub-plots (a-c) display the probability distribution functions for three key topological properties for the US Pacific Northwest: Average Lifetime of holes, Betti curve-based features of holes, and Bottleneck amplitude of holes. These distributions reveal a striking similarity between sliding-type and complex landslide failures.} Sub-plots (d-e) show the probability of complex landslides belonging to each of the failure types class (slide, flow, and fall) as predicted by the model. Box plot (d) shows the complex landslide samples that occur from "Translational rock slides followed by rock falls" as documented in the Statewide Landslide Information Database for Oregon (SLIDO). Model predictions indicate sliding {failure} to be the predominant type of failure, which are most likely translational slides according to SLIDO.
Similarly, box plot (e) shows the complex landslide samples that occur from "Rotational slides followed by flows" as recorded in SLIDO. Model predictions indicate slide {type} to be the predominant type of failure, which most likely ruptures rotationally. Beige bars illustrate sliding mechanisms while bars with darker and lighter shades of blue illustrate flows and fall {types}, respectively. Note the number of annotated complex failures with behavioral definitions by SLIDO in box plot (a) constitutes 198 samples and box plot (b) constitutes 230 samples.}

% The persistence diagram captures the birth and death of structures (such as holes and connected components), and using this information, we can calculate the topological properties of the landslide's shape.

\label{fig:complex_plot}
\end{figure*}

% For example, failure mechanisms that are very different (slides vs flows) might share the same area geometrically but will have different shapes overall, slides being more compact than the usual elliptical/elongated flows.
Therefore, to assess and evaluate the differences between the classical and topological properties, we compare them in this section. This comparison was based on KDE plots that represent the PDFs of the samples for each failure {type}. We also plotted box plots to compare the median values and distribution of said values between the geometric and topological properties. As we see in Figure~\ref{fig:geom_topo}-b, the PDFs of the failure types are very similar to each other, specifically when looking at the ellipticity, semi-major axis, perimeter, and width. However, when comparing them to the topological properties Figure~\ref{fig:geom_topo}-a, we observe that the PDFs of the failure types are more dissimilar to each other under each property (e.g., the average lifetime of holes, bottleneck amplitude of holes, Wasserstein amplitude of holes). This can be the reason why the random forest models show promising results, as the PDFs are dissimilar enough to find evident differences between each failure type when using the topological properties/features.

%\begin{figure*}[t!]
%\includegraphics[width=\columnwidth]{box_plot_geom_tda.pdf}\hfill
%\caption{The figure shows the probability distribution functions of the geometrical and the topological features for each failure mechanisms type-- slides, flows, complex, and falls. The y-axis shows the probability density values (calculated using kernel density estimation), and the x-axis shows the value of topological or geometrical attributes. The Geometric features from top to bottom are- area ($A$), perimeter ($P$), ratio of area to perimeter $\frac{A}{P}$, convex hull-based measure ($C_h$), minor($s_m$), width ($W$) of the minimum area bounding box fitted to the polygon.}
%\label{fig:geom_topo}
%\end{figure*}

\section{Other measures to evaluate model
performance}

In order to evaluate the performance of the method, we also calculated the confusion matrix and other accuracy metrics like the True Positive Rate, True Negative Rate, False Positive Rate, False Negative Rate, and the F1-score. 

The True positive rate (also known as Sensitivity, Recall) (equation 1) and true negative rate (also known as Specificity) (equation 2) are performance metrics that are used to assess a model's accuracy in accurately detecting positive and negative instances. The number of genuinely negative instances identified as positive by the model are known as false positives (FP) (equation 3). The number of cases that are genuinely positive but are categorized as negative by the model are known as false negatives (FN) (equation 4). 

\begin{align}
TPR\text{ }(= Recall) &= \frac{True\ Positives}{True\ Positives + False\ Negatives}  \label{eq:tpr} \\
TNR &= \frac{True\ Negatives}{True\ Negatives + False\ Positives} \label{eq:tnr} \\
FPR &= \frac{False\ Positives}{True\ Negatives + False\ Positives} \label{eq:fpr} \\
FNR &= \frac{False\ Negatives}{True\ Positives + False\ Negatives} \label{eq:fnr}
\end{align}

The F1-score (equation 6) is the harmonic mean of precision (equation 5) and recall (equation 1), and it is used to balance the precision-to-recall trade-off. Precision is the number of correct positive predictions produced by the model out of all correct positive predictions made by the model, and recall is the number of correct positive predictions made out of all correct positive occurrences.

\begin{align}
Precision &= \frac{True\ Positives}{True\ Positives + False\ Positives} \label{eq:precision} \\
F1\text{-}score &= \text{ } 2 \cdot \frac{Precision \cdot Recall}{Precision + Recall} \label{eq:f1score}
\end{align}

In Figure~\ref{fig:confusion}, we see the confusion matrix and the respective scores of the TPR, TNR, FPR, FNR, and F1-score of both Italy and the US Pacific Northwest.

%%===================================================%%
%% For presentation purpose, we have included        %%
%% \bigskip command. please ignore this.             %%
%%===================================================%%

%%===========================================================================================%%
%% If you are submitting to one of the Nature Portfolio journals, using the eJP submission   %%
%% system, please include the references within the manuscript file itself. You may do this  %%
%% by copying the reference list from your .bbl file, paste it into the main manuscript .tex %%
%% file, and delete the associated \verb+\bibliography+ commands.                            %%
%%===========================================================================================%%
%% if required, the content of .bbl file can be included here once bbl is generated
%%\input sn-article.bbl
%% Default %%
%%\input sn-sample-bib.tex%

% \begin{figure*}[t!]
% \includegraphics[width=\columnwidth]{tda_features_italy.pdf}\hfill
% \caption{The figure shows the landslide failure type samples for slide, flow, complex and fall.}
% \label{fig:datafig1}
% \end{figure*}

% \begin{figure*}[t!]
% \includegraphics[width=\columnwidth]{tda_features_usa.pdf}\hfill
% \caption{The figure shows the landslide failure type samples for slide, flow, complex and fall.}
% \label{fig:datafig1}
% \end{figure*}

% \begin{figure*}[t!]
% \includegraphics[width=\columnwidth]{comine_usa_italy.pdf}\hfill
% \caption{The figure shows the landslide failure type samples for slide, flow, complex and fall.}
% \label{fig:datafig1}
% \end{figure*}

\bibliographystyle{unsrt}  
\bibliography{references}

\end{document}